\documentclass[runningheads]{llncs}
\usepackage[english]{babel}
\usepackage{caption} 
\usepackage{graphicx}
\usepackage{amsmath}
\usepackage{algpseudocode}
\usepackage[boxed,vlined, linesnumbered]{algorithm2e}
\DontPrintSemicolon
\usepackage{import}
\usepackage{amssymb}
\usepackage{multirow}
\usepackage{bussproofs}
\usepackage{xcolor}
\usepackage{rotating}
\usepackage{tabularx}
\usepackage{cleveref}
\usepackage{todonotes}
\usepackage{makecell}
\usepackage{subcaption}
\usepackage{todonotes}

\begin{document}
\title{When GNNs Met a Word Equations Solver: Learning to Rank Equations}

\titlerunning{When GNNs Met a Word Equations Solver}

\newcommand{\answerTodo}[1]{\todo[color=green!40]{#1}} 
\newcommand*\samethanks[1][\value{footnote}]{\footnotemark[#1]}

\newcommand\SAT{\ensuremath{\mathrm{SAT}}\xspace}
\newcommand\UNSAT{\ensuremath{\mathrm{UNSAT}}\xspace}
\newcommand\UNKNOWN{\ensuremath{\mathrm{UNKNOWN}}\xspace}
\newcommand\BT{\ensuremath{\mathit{BT}}\xspace}
\newcommand\OURSOLVER{\textsf{DragonLi}\xspace}

\newcommand\myCall[2]{\ensuremath{\mathit{#1}(#2)}}

\SetKwRepeat{Do}{do}{while}

 \author{
   Parosh Aziz Abdulla
   \inst{1} \orcidID{0000-0001-6832-6611}
   \and 
   Mohamed Faouzi Atig  
    \inst{1} 
 	\and
 	Julie Cailler 
 	\inst{3} \orcidID{0000-0002-6665-8089}
 	\and
 	Chencheng Liang
 	\inst{1} \orcidID{0000-0002-4926-8089}
    \and
 	Philipp R{\"u}mmer
 	\inst{1,2} \orcidID{0000-0002-2733-7098}
 }
 \authorrunning{P. Abdulla et al.}

 %
 \institute{
 Uppsala University, Uppsala, Sweden\\
 \and
 University of Regensburg, Regensburg, Germany\\
 \and
 University of Lorraine, CNRS, Inria, LORIA, Nancy, France\\
 }

\maketitle              

\begin{abstract}
Nielsen transformation is a standard approach for solving word equations:
by repeatedly splitting equations and applying simplification steps,
equations are rewritten until a solution is reached.
When solving a conjunction of word equations in this way, the performance
of the solver will depend considerably on the order in which equations
are processed.
In this work, the use of Graph Neural Networks (GNNs)
for ranking word equations before and during the solving process is explored.
For this, a novel graph-based representation for word equations is presented, preserving global information across conjuncts, enabling the GNN to have a holistic view during ranking.
To handle the variable number of conjuncts, three approaches to adapt a multi-classification task to the problem of ranking equations are proposed.
The training of the GNN is done with the help of minimum unsatisfiable subsets (MUSes) of word equations.
The experimental results show that, compared to state-of-the-art string solvers, the new framework solves more problems in benchmarks where each variable appears at most once in each equation. 

\end{abstract}

\section{Introduction}

%

A \textit{word equation} is an equality between two \textit{strings} that may contain variables representing unknown substrings. Solving a \textit{word equation problem} involves finding assignments to these variables that satisfy the equality. Word equations are crucial in string constraints encountered in program verification tasks, such as validating user inputs, ensuring proper string manipulations, and detecting potential security vulnerabilities like injection attacks.
The word equation problem is decidable, as shown by Makanin~\cite{Mak77}; while the precise complexity of the problem is still open, it is know to be NP-hard and in PSPACE~\cite{814622}.

Abdulla et al.~\cite{10.1007/978-3-031-78709-6_14} recently proposed a Nielsen transformation-based algorithm for solving word equation problems~\cite{Nielsen1917}.  
This algorithm solves word equations by recursively applying a set of inference rules to branch and simplify the problem until a solution is reached, in a tableau-like fashion.
When multiple word equations are present, the algorithm must select the equation to process next at each proof step. 
This selection process is critical and heavily influences the performance of
the algorithm, as the unsatisfiability of a set of equations can often be shown by identifying a small unsatisfiable core of equations. At the same time,
the search tree can contain infinite branches on which no
solutions can be found, so that bad decisions can lead a solver astray.
The situation is similar to the case of first-order logic theorem provers, where the choice of clauses to process plays a decisive role in determining efficiency.
In the latter context, several deep learning techniques have been introduced to guide Automated Theorem Provers (ATPs)~\cite{Jakubuv2020,BartekSuda2021,10.1007/978-3-030-79876-5_31,abdelaziz2021learningguidesaturationbasedtheorem,LPAR2023:How_Much_Should_This}.  
However, for word equation problems, the application of learning techniques for selecting equations remains largely unexplored.



In this work, we employ Graph Neural Networks (GNNs)~\cite{DBLP:journals/corr/abs-1806-01261} to guide the selection of word equations at each iteration of the algorithm. Our research complements existing techniques for learning branching heuristics in word equation solvers~\cite{10.1007/978-3-031-78709-6_14} (Appendix~\ref{appendix:relation-to-previous-paper}).
We refer to the selection step as the \emph{ranking process}.
For this, we enhanced the existing algorithm~\cite{10.1007/978-3-031-78709-6_14} to enable the re-ordering of conjunctive word equations.
The extension preserves the soundness and the completeness (for finding solutions) of the algorithm.
We refer to this extended algorithm as the \emph{split algorithm} throughout the paper.

The primary challenge in training a deep learning model to guide the ranking process lies in managing a variable number of inputs. In our work, this specifically involves handling a varying number of word equations depending on the input.
Unlike with branching heuristics, which have to handle only a fixed and small number of branches (typically 2 to 3), the ranking process must handle a variable number of conjuncts.
To address this challenge, we adapt multi-classification models to accommodate inputs of varying sizes using three distinct approaches.
Additionally, to effectively train the GNNs, we enhance the graph representations of word equations from \cite{10.1007/978-3-031-78709-6_14} by incorporating global term occurrence information.

Our model is trained using data from two sources: (1)~Minimal Unsatisfiable Subsets (MUSes) of word equations computed by other solvers, and (2)~data extracted by running the split algorithm with non-GNN-based ranking heuristics.
MUSes computed by solvers such as \textsf{Z3}~\cite{demoura2008z} and \textsf{cvc5}~\cite{10.1007/978-3-030-99524-9_24} help detect unsatisfiable conjuncts early, enabling prompt termination and improved efficiency.
When the split algorithm tackles conjunctive word equations, each ranking decision creates a branch in a decision tree. By extracting the shortest path from this tree, we obtain the most effective sequence of choices, which we then use as  training data.


Moreover, we explore seven options that combine the trained model with both random and manually designed heuristics for the ranking process.


We evaluated our framework on artificially generated benchmarks inspired by \cite{day2019solving}.
The benchmarks are divided into two categories: \emph{linear} and \emph{non-linear}, where linear means that, within a single equation, a variable can occur only once, while non-linear allows a variable to appear multiple times.
Note that this definition of linearity applies to individual equations: in systems with multiple equations, even if each equation is linear, shared variables can cause a variable to appear multiple times within the system.

Finally, we compare our framework with several leading SMT solvers and a word equation solver, including \textsf{Z3}, \textsf{Z3-Noodler}~\cite{10.1007/978-3-031-57246-3_2}, \textsf{cvc5}, \textsf{Ostrich}~\cite{DBLP:journals/pacmpl/ChenHLRW19}, and \textsf{Woorpje}~\cite{day2019solving}.
The experimental results show that for linear problems, our framework outperforms all leading solvers in terms of the number of solved problems.
For non-linear problems, when the occurrence frequency of the same variables (non-linearity) is low, our algorithm remains competitive with other solvers. 

In summary, the contributions of this paper are as follows:
\begin{itemize}
	\item We adapt the Nielsen transformation-based algorithm~\cite{10.1007/978-3-031-78709-6_14} to allow control over the ordering of word equations at each iteration.
	\item We develop a framework to train and deploy a deep learning model for ranking and ordering conjunctive word equations within the split algorithm. The model leverages MUSes generated by leading solvers and uses graph representations enriched with global information of the formula. We propose three strategies to adapt multi-classification models for ranking tasks and explore various integration methods within the split algorithm.
	\item Experimental results demonstrate that our framework performs effectively on linear problems, with the deep learning model significantly enhancing performance. However, its effectiveness on non-linear problems is constrained by the limitations of the inference rules.
\end{itemize}


\section{Preliminaries}\label{section:preliminaries}

We first define the syntax of word equations and the concept of satisfiability. Next, we explain the message-passing mechanism of Graph Neural Networks (GNNs) and describe the specific GNN model employed in our experiments.

\subsubsection{Word Equations.}
\label{section:preliminaries:word-equations}

We assume a finite non-empty alphabet $\Sigma$ and write
$\Sigma^{*}$ for the set of all strings (or words) over $\Sigma$.
The empty string is denoted by $\epsilon$.
We work with a set $\Gamma$ of string variables, ranging over words in $\Sigma^{*}$. The symbol $\cdot$ denotes the concatenation of two strings; in our examples, we often write~$uv$ as shorthand for $u \cdot v$.
The syntax of word equations is defined as follows:
\begin{align*}
  \text{Formulae} ~ \phi & ::= \mathit{true} \mid e \land \phi &
 \text{Words} ~ w & ::= \epsilon \mid t \cdot w   \\
 \text{Equations} ~ e & ::= w = w   &
          \text{Terms} ~ t & ::= X \mid c  
\end{align*}
where $X \in \Gamma$ ranges over variables and $c \in \Sigma$ over letters.

%

\begin{definition}[Satisfiability of conjunctive word equations]
  A formula~$\phi$ is \emph{satisfiable} \emph{(\SAT)} if there exists a substitution~$\pi: \Gamma \rightarrow \Sigma^{*}$ such that, when each variable $X\in \Gamma $ in $\phi$ is replaced by $\pi(X)$, all equations in $\phi$ hold.
\end{definition}

\begin{definition}[Linearity of a word equation]
	A word equation is called \emph{linear} if each variable occurs at most once. Otherwise, it is \emph{non-linear}.
\end{definition}

\subsubsection{Graph Neural Networks.}

%


\emph{Message Passing-based GNNs} (MP-GNNs) \cite{DBLP:journals/corr/GilmerSRVD17}
are designed to learn features of graph nodes (and potentially the entire graph) by iteratively aggregating and transforming feature information from the neighborhood of a node.
Consider a graph $ G = (V, E) $, with $ V $ as the set of nodes and $ E \subseteq V \times V $ as the set of edges. Each node $ v \in V $ has an initial representation $ x_{v} \in \mathbb{R}^{n}$ and a set of neighbors $ N_{v} \subseteq V $.
In an MP-GNN comprising $ T $ message-passing steps, node representations are iteratively updated. The initial node representation of $v$ at time step 0 is $H_{v}^{0} = x_{v}$. At each step $ t $, the representation of node $ v $, denoted as $ H_{v}^{t} $, is updated using the equation:
\begin{equation}\label{eq:MPGNN}
H_{v}^{t} = \eta_{t}(\rho_{t}(\{H_{u}^{t-1} \mid u \in N_{v}\}), H_{v}^{t-1}),
\end{equation}
where $H_{u}^{t-1}$ is the node representation of $u$ in the previous iteration $t-1$, and node $u$ is a neighbor of node $v$.
In this context, $\rho_{t}:(\mathbb{R}^{n})^{|N_{v}|} \rightarrow \mathbb{R}^{n}$ is an aggregation function with trainable parameters (e.g., an MLP followed by sum, mean, min, or max) that aggregates the node representations of $v$'s neighboring nodes at the $t$-th iteration.
Along with this, $\eta_{t}:(\mathbb{R}^{n})^2\rightarrow \mathbb{R}^{n}$ is an update function with trainable parameters (e.g., an MLP) that takes the aggregated node representation from $\rho_{t}$ and the node representation of $v$ in the previous iteration as input, and outputs the updated node representation of $v$ at the $t$-th iteration.


In this study, we employ \emph{Graph Convolutional Networks} (GCNs)~\cite{kipf2017semisupervised} to guide our algorithm due to their computational efficiency to generalize across tasks without the need for task-specific architectural modifications.
In GCNs, the node representation $H_{v}^{t}$ of $v$ at step $t \in \{1, ..., T\}$ where $T \in \mathbb{N}$ is computed by 
\begin{equation}\label{eq:GCN}
 H_{v}^{t} = \text{ReLU}(\text{MLP}^{t}(\text{mean}\{H_{u}^{t-1} \mid u \in N_{v} \cup \{v\}\})),
\end{equation}
where each $\text{MLP}^{t}$ is a fully connected neural network,  ReLU (Rectified Linear Unit)~\cite{agarap2018deep} is the non-linear function $f(x)=max(0,x)$, and $H_{v}^{0}=x_{v}$.

\section{Split Algorithm with Ranking}
\label{section:split-algorithm}

%
%
%
%


\subsubsection{Split Algorithm.} Algorithm~\ref{algorithm:splitEquation}, \(\textsc{splitEquations}\), determines the satisfiability of a word equation formula \(\phi\) by recursively applying the inference rules from \cite{10.1007/978-3-031-78709-6_14}.
The inference rules are provided in Appendix~\ref{app:split-rules} for a convenient reference.

The algorithm begins by 
checking the
satisfiability of the conjunctive formula (Line~\ref{alg:simplifyAndCheckFormula}).
If all word equations can be eliminated in this way,  then \(\phi\) is \SAT. 
If any conjunct is unsatisfiable (\UNSAT), then \(\phi\) is \UNSAT. Otherwise, the satisfiability status remains \emph{unknown} (UKN). If $\phi$ is in one of the first two cases, its status is returned (Line~\ref{alg:neqUnknown}).

Otherwise (Line~\ref{alg:eqUnknown}), \(\textsc{RankEqs}\) orders all
conjuncts using either manually designed or data-driven methods. Next,
the function \(\textsc{ApplyRules}\) matches and applies the
corresponding inference rules to generate branches—alternative
prospective solving paths for the same equation. This step is called
the \emph{branching process}. Notably, rules $R_7$ and $R_8$ generate
two and three branches, respectively, while all the other rules do not
cause any branching.

Next, the $\textsc{splitEquations}$ call (Line~\ref{alg:call-splitEquations}) recursively checks the satisfiability of each branch. Let $\{ b_{1}, \dots, b_{n} \}$ be the set of branches. $\phi$ has status \SAT if at least one branch $b_{i}$ is satisfiable, \UNSAT if all branches are unsatisfiable, and UKN otherwise.

%
%

\begin{algorithm}[t]
	\caption{\textsc{SplitEquations} algorithm.}\label{algorithm:splitEquation}
	\SetKwFunction{splitEquations}{splitEquations}
	\KwData{A formula $\phi$}
	\KwResult{The satisfiability status of $\phi$ (i.e., \textit{SAT}, \textit{UNSAT}, or \textit{UKN}) and the simplified version of $\phi$}
	
	\Begin{
	
		$\mathit{res} \leftarrow $ \Call{CheckFormulaSatisfiability}{$\phi$}\label{alg:simplifyAndCheckFormula}
		
		\lIf{res $\neq$ UKN}{\label{alg:neqUnknown}
			\KwRet res, $\phi$
		}
		\Else {\label{alg:eqUnknown}
			$\phi_s$ = \Call{RankEqs}{$\phi$} \tcp*{Ranking process}  \label{alg:rankEq}
			
			\textit{Branches} = \Call{applyRules}{$\phi_s$} \tcp*{Branching process}

			$\mathit{uknFlag} \leftarrow 0$
			
			\For{b in Branches}{ \label{alg:branchEq}
				$\mathit{res_b}, \phi_b$=\Call{splitEquations}{\textit{b}}\label{alg:call-splitEquations}
				
				\lIf{$\mathit{res_b}$ = \textit{SAT}}{
					\KwRet \textit{SAT}, $\phi_b$
				}
				
				\lIf{$\mathit{res_b}$ = \textit{UKN}}{
					$\mathit{uknFlag} \leftarrow 1$
				}
			}
			
			\lIf{$\mathit{uknFlag} = 1$}{
				\KwRet \textit{UKN}, $\phi$
			}
			\lElse {
				\KwRet \textit{UNSAT}, $\phi$ \label{alg:returnUnsat}
			}
			
		}
	}
\end{algorithm}


Since the inference rules apply to the leftmost equation, the performance and termination of the algorithm are strongly influenced by both the order in which branches are processed (Line~\ref{alg:branchEq}) and the ordering of equations in \(\phi\) (Line~\ref{alg:rankEq}). While the impact of branch ordering has been studied in~\cite{10.1007/978-3-031-78709-6_14}, this paper explores whether employing a data-driven heuristic in \(\textsc{RankEqs}\) can enhance termination.

The baseline option to implement $\textsc{RankEqs}$ is referred to as
\textbf{RE1: Baseline}. It computes the priority of a word equation
$p$ using the following definition:
\begin{align*}
	p &~=~
	\begin{cases}
		1 & \text{if $\epsilon = \epsilon$
		} 
		\\
		2 & \text{otherwise, if $\epsilon = u\cdot v$ or $u\cdot v=\epsilon$ 
			} 
		\\
		3 & \text{otherwise, if $a\cdot u=b\cdot v$ or $u \cdot a=v \cdot b$ 
			} 
		\\
		4 & \text{otherwise, if $a\cdot u=a \cdot v$ 
		} 
		\\
		5 & \text{otherwise}
	\end{cases}
\end{align*}
where $a, b \in \Sigma$, and $u, v$ are sequences of variables and letters.
Smaller numbers indicate higher priority, assigning greater precedence to simpler cases where satisfiability is obvious.  
Word equations with the same priorities between 1 and 4 are further ordered by their length (i.e., the number of terms), with shorter equations taking precedence.  
For word equations with a priority of 5, the original input order is maintained.
The newly created equations inherit the ranking of their parents.  
We refer to the split algorithm using \textbf{RE1} for \(\textsc{RankEqs}\) as \textsf{DragonLi}.  
The correctness of Algorithm~\ref{algorithm:splitEquation} follows directly from the soundness and local completeness of the inference rules in \cite{10.1007/978-3-031-78709-6_14}:

\begin{lemma}[Soundness of Algorithm~\ref{algorithm:splitEquation}]
	For a conjunctive word equation formula $\phi$, if Algorithm~\ref{algorithm:splitEquation} terminates with the result \SAT or \UNSAT, then $\phi$ is \SAT or \UNSAT, respectively. 
\end{lemma}

%
%

\subsubsection{AND-OR Tree.}  
\begin{figure}[h]
	\fbox{\includegraphics[width=\textwidth-2\fboxsep,trim=40 10 60 20]{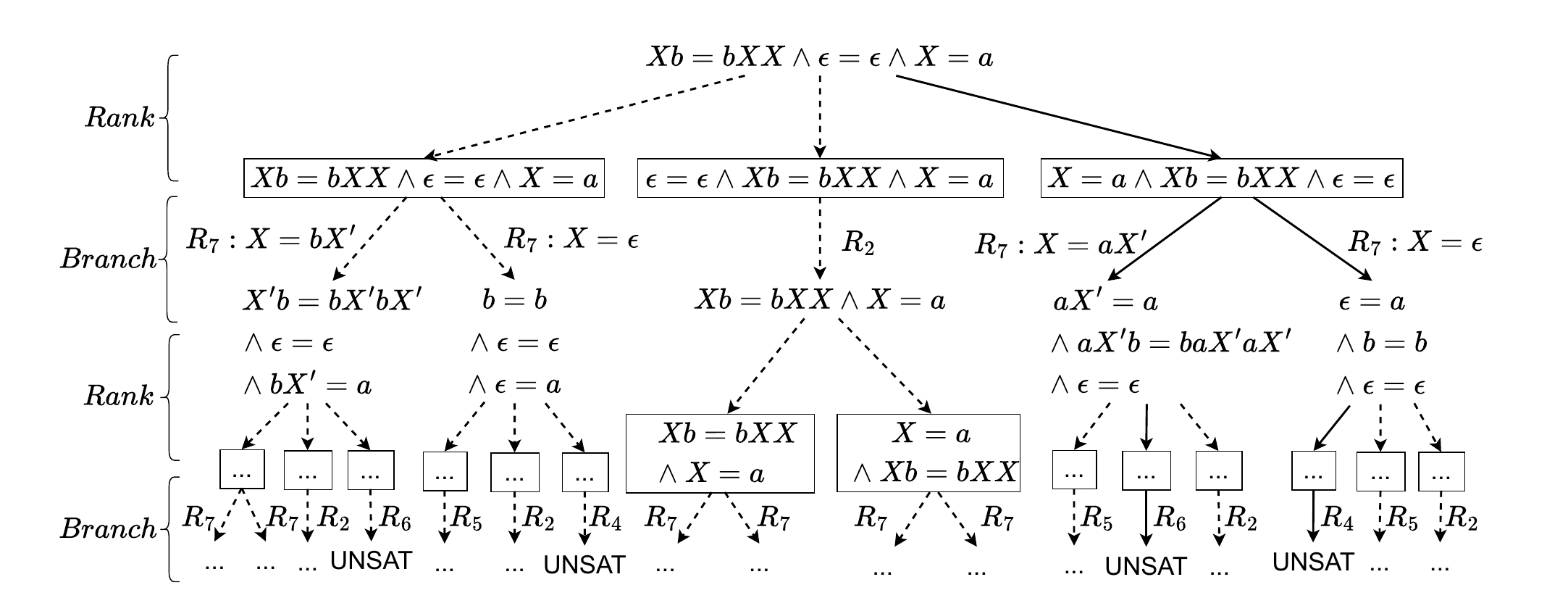}}
	\caption{AND-OR tree resulting from the word equation $Xb = bXX \wedge \epsilon = \epsilon \wedge X = a$. The formulas enclosed in boxes are generated by $\textsc{RankEqs}$, while the formulas without boxes are obtained from $\textsc{ApplyRules}$.
	}
	\label{figure:proof-tree}
\end{figure}

The search tree explored by the algorithm can be represented as an
AND-OR tree, as shown in Figure~\ref{figure:proof-tree}. 
The example illustrates the three paths, each placing different equations in the first position,
generated by the ranking and branching process to solve the word equation \(\phi = (Xb = bXX \wedge \epsilon = \epsilon \wedge X = a)\), where \(a, b \in \Sigma\) and \(X \in \Gamma\).  

\begin{example}  
	In the first step, \(\phi\) can be reordered in three distinct
        ways by prioritizing one conjunct to occupy the leftmost
        position (we ignore the order of the rest two equations, as
        their order does not influence the next rule application). 
	Thus, the root of the tree branches into three paths.  
	For each ranked formula, the inference rules are then applied to execute the branching process.  
	By iterating these two steps alternately, the complete AND-OR tree is constructed.  
	Notably, continuously selecting the leftmost branch that prioritizes \(Xb = bXX\) at the root and applying the left branch of \(R_{7}\) may lead to non-termination, as the length of the word equation keeps increasing.  
	In contrast, prioritizing \(X = a\) at the root results in a solution (\UNSAT) at a relatively shallow depth, avoiding the risk of non-termination caused by further ranking and branching.  
	In this case, exploring only a single branch during the ranking process suffices to determine the satisfiability of \(\phi\).  
	This optimal path is highlighted with solid edges. 
\end{example}

\section{Guiding the Split Algorithm}
\label{section:learning}
This section details the training and application of a GNN model in Algorithm~\ref{algorithm:splitEquation}.
We first describe the process of collecting training data, followed by the graph-based representation of each word equation. Next, we outline three model structures for ranking a set of word equations. Finally, we discuss methods for integrating the trained model back into the algorithm.

\subsection{Training Data Collection}
\label{section:training_data_collection}

Assume that $\phi$ is an unsatisfiable conjunctive word equation consisting of a set of conjuncts $\mathcal{E}$.

\begin{definition}[Minimal Unsatisfiable Set]
	\label{def:MUS}
	A subset $U \subseteq \mathcal{E}$ is a \emph{Minimal Unsatisfiable Set} (MUS) if the conjunction of $U$ is unsatisfiable, and for all conjuncts $e \in U$, the conjunction of subset $U \setminus \{e\}$ is satisfiable.
\end{definition}

We collect training data from two sources:  
(1)~MUSes extracted by other solvers, including \textsf{Z3}, \textsf{Z3-Noodler}, \textsf{cvc5}, and \textsf{Ostrich}; and  
(2)~formulas from the ranking process that lie on the shortest path from the subtree
leading to \UNSAT\ in the AND-OR trees.  
A numerical example of these two sources is provided in Section~\ref{section:experiment-train-data-collection}.  

For training data from source (1), we first pass all problems to \textsf{DragonLi}.  
Next, we identify unsolvable problems and forward them to other solvers.  
If any solver successfully solves a problem, we select the one that finds a solution in the shortest time.  
This solver is then used to extract the MUS by exhaustively checking the satisfiability of all subsets of the conjuncts.  
Finally, each conjunct within a set of word equations is labeled based on its membership in the MUS and its length.

Formally, given a formula $\phi = e_{1} \wedge \dots \wedge e_{n}$, its conjuncts are denoted $\mathcal{E} = \{e_{1}, \dots, e_{n}\}$, and an MUS $U \subseteq \mathcal{E}$. The corresponding labels of $e_i \in \mathcal{E}$ are $Y_{n} = \{ y_{1}, \dots, y_{n} \}$, where $y_i \in \{0, 1\}$, and their length is denoted $|e_{i}|$.
The label $y_{i}$ is computed as follows:
\begin{align}
	y_i &=
	\begin{cases}
		1 & \text{if } e_{i} \in U \text{ and } |e_{i}| = \min \big( \{ |e| \mid e \in U \} \big), \\
		0 & \text{otherwise}.
	\end{cases}
\end{align}
We assign label 1 only to the shortest equation in the MUS, rather than labeling all MUS equations as 1 and non-MUS equations as 0, because the algorithm selects only one equation to proceed at each iteration. Our goal is to identify the most efficient choice. We assume that the shortest equation in the MUS is more likely to lead to quicker termination, as the branching process aims to reduce equation length until a form is reached where satisfiability (or unsatisfiability) can be easily concluded.


To collect training data from source (2), we pass the problems, along with the MUS extracted from other solvers, to \textsf{DragonLi}.  
If \textsf{DragonLi} solves the problem, multiple paths to \UNSAT are generated by sequentially prioritizing each equation at the leftmost position in the ranked word equation.  

Subsequently, we export and label each conjunctive word equation along the shortest path in the subtree
leading to \UNSAT. Formally, given a set of conjuncts~$\mathcal{E}=\{e_{1}, \dots, e_{n}\}$ of a conjunctive word equation, the corresponding labels $Y_{n} = (y_{1}, \dots, y_{n})$ are computed by 
\begin{align}
	y_i &=
	\begin{cases}
		1 & \text{if } e_{i}~\text{in the shortest path of a subtree leading to \UNSAT}, \\
		0 & \text{otherwise}.
	\end{cases}
\end{align}

For both sources, when $\sum_{i=1}^{n}y_i >1$, we keep the first equation with label 1 and  discard the rest equations with label 1 
to ensure $\sum_{i=1}^{n}y_i =1$.
When $\sum_{i=1}^{n}y_i = 0$, we discard this training data due to no positive label.

\subsection{Graph Representation for Conjunctive Word Equations}
\label{subsection:graph-representation}
The graph representation of a single word equation is discussed in \cite{10.1007/978-3-031-78709-6_14}. However, since word equations are interconnected through shared variables, ranking them requires not only local information about individual equations but also a global perspective. By considering the entire set of word equations collectively, we can incorporate dependencies and shared structures, improving the ranking process.

To achieve this, we first represent each conjunctive word equation independently. Then, we compute the occurrences of variables and letters across all equations and integrate this global information into each individual graph representation. This enriched representation captures both the complexity of individual equations and their interactions within the system.

In details, the graph representation of a word equation is defined as 
$G = (V, E, v_{=}, V_{\text{T}}, V_{\text{Var}}, V_{\text{T}}^{0}, V_{\text{T}}^{1}, V_{\text{Var}}^{0}, V_{\text{Var}}^{1})$, 
where $V$ is the set of nodes, $E \subseteq V \times V$ is the set of edges, and $v_{=} \in V$ is a special node representing the ``='' symbol. The sets $V_{\text{T}} \subseteq V$ and $V_{\text{Var}} \subseteq V$ contain letter and variable nodes, respectively. Additionally, $V_{\text{T}}^{0}$ and $V_{\text{T}}^{1}$ are special nodes representing letter occurrences and $V_{\text{Var}}^{0}$ and $V_{\text{Var}}^{1}$ analogously represent variable occurrences.

Figure~\ref{fig:word-equation-graphs} illustrates the two steps involved in constructing the graph representation of the conjunctive word equations $XaX = Y \wedge aaa = XaY$, where $\{X, Y\} \subseteq \Gamma$ and $a \in \Sigma$:

\begin{figure}[t]
  \begin{center}
		\includegraphics[width=0.8\textwidth,trim=20 15 25 10]{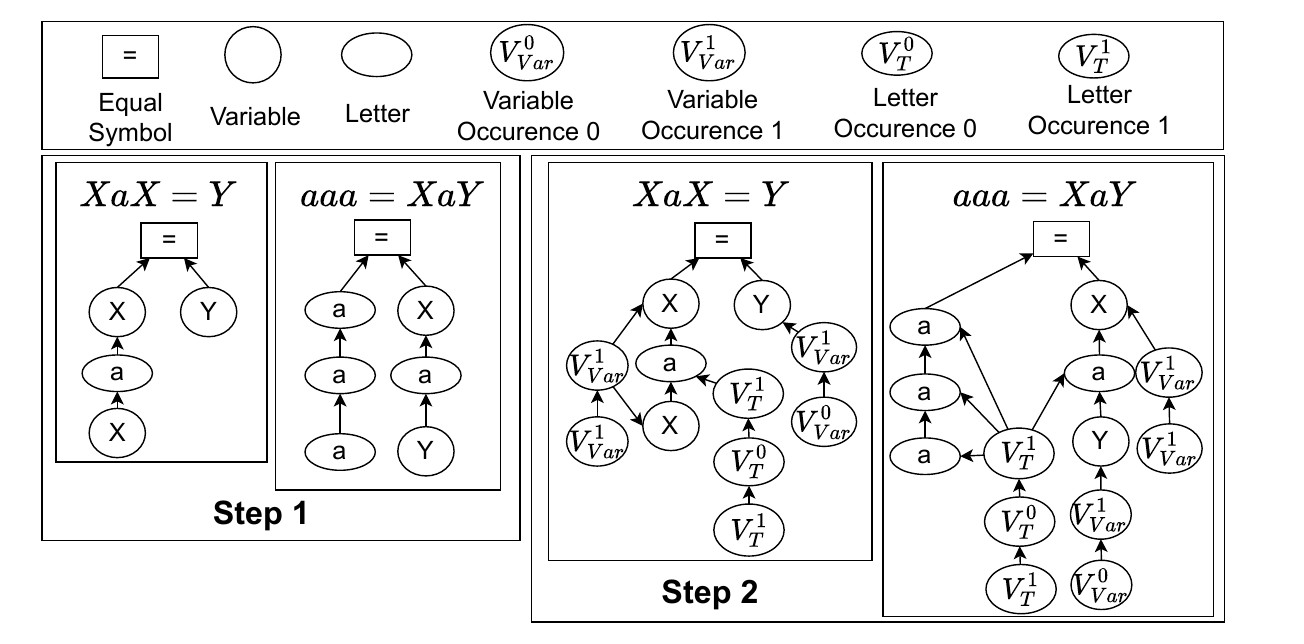}
  \end{center}

  \caption{The steps for constructing graph representation for the conjunctive word equations $XaX=Y \wedge aaa=XaY$ where $X, Y$ are variables and $a$ is a letter.
  }\label{fig:word-equation-graphs}
\end{figure}

\begin{itemize}
	\item \textbf{Step 1:} Inspired by Abstract Syntax Trees (ASTs), we begin to build the graph by placing the ``$=$'' symbol as the root node. The left and right children of the root represent the leftmost terms of each side of the equation, respectively. Subsequent terms are organized as singly linked lists of nodes.
	
	\item \textbf{Step 2:} Calculate the number of occurrences of all terms across the conjunctive word equations. In this example, $\text{Occurrence}(X) = 3$, $\text{Occurrence}(Y) = 2$, and $\text{Occurrence}(a) = 5$. Their binary encodings are $11$, $10$ and $101$ respectively. We encode these as sequentially connected nodes: $(V_{Var}^{1}, V_{Var}^{1})$ for $X$, $(V_{Var}^{1}, V_{Var}^{0})$ for $Y$, and $(V_{T}^{1}, V_{T}^{0}, V_{T}^{1})$ for $a$. Finally, we connect the roots of these nodes to their corresponding variable and letter nodes.
\end{itemize}

We chose binary encoding because using unary encoding would significantly increase the graph size, making computation inefficient. Higher-base encodings like ternary or decimal tend to blur structural distinctions that means different values may be represented using the same number of nodes, making it hard for the graph structure to reflect meaningful differences.  Binary encoding strikes a balance. It keeps the graph size manageable while preserving enough structural information for our GNN to effectively process word equations at the scale we target.

The rationale behind our other choices of graph representation for word equations, along with a discussion of alternative representations, is provided in Appendix~\ref{appendix:word-equation-graph-encoding}.

\subsection{Training of Graph Neural Networks}

In the function \textsc{RankEqs} of Algorithm~\ref{algorithm:splitEquation}, equations can be ranked and sorted based on predicted rank scores from a trained model. 
Given a conjunctive word equation $\phi = e_1 \wedge \dots \wedge e_n$, the model outputs a \textit{ranking}, i.e., a list of real numbers $\hat{Y}_{n} = (\hat{y}_1, \dots, \hat{y}_n)$ in which a higher value indicates a higher rank. 
For example, for a conjunctive word equation $e_1 \wedge e_2$, the model might output $\hat{Y}_{2} = (0.3, 0.7)$, indicating that $e_2$ is expected to lead to a solution more quickly than $e_1$, and the equations should be reordered as $e_2 \wedge e_1$.

\subsubsection{Forward Propagation.}

To compute this ranking, we first transform the word equations $\{e_{1},...,e_{n}\}$ to their graph representations $G= \{G_{1},...,G_{n}\}$ where $G_{i}=(V, E, v_{=}, V_{\text{T}}, V_{\text{Var}}, V_{\text{T}}^{0}, V_{\text{T}}^{1}, V_{\text{Var}}^{0}, V_{\text{Var}}^{1})$.
Each node $v\in V$ is first assigned an integer representing the node
type: $v\in \bigcup \{V_{T}, V_{Var}, V_{T}^{0}, V_{T}^{1}, V_{Var}^{0}, V_{Var}^{1}\}\cup\{v_{=}\}$.
Those integers are then passed to a trainable embedding function $\text{MLP}_{0}:\mathbb{Z} \rightarrow \mathbb{R}^{m}$ to compute all the initial node representations $H_{i}^{0}$ in $G_{i}$.


Equation~\eqref{eq:GCN} defines how node representations are updated. By iterating this update rule, we obtain the node representations \( H_i^{t} = \text{GCN}(H_i^{t-1}, E) \) for \( t \in \{1, \ldots, T\} \), where the relation \( E \) is used to identify neighbors. Subsequently, the representation of the entire graph is obtained by summing the node representations at time step \( T \), resulting in $ H_{G_i} = \frac{1}{n}\sum_{i=1}^{n} H_i^{T} $.

Then, we introduce three ways to compute the $\hat{Y}_{n}$:
\begin{itemize}
	\item \textbf{Task 1}: Each graph representation $H_{G_{i}}$ is given to a trainable classifier 
	$\text{MLP}_{1} : \mathbb{R}^{m} \rightarrow \mathbb{R}^{2}$, which outputs 
	$\mathbf{z}_{i}=\text{MLP}_{1}(H_{G_{i}}) = (z_{1}, z_{2})$. The score for graph $i$ is then computed as
	$y_{i} = \textit{softmax}(\mathbf{z}_{i})_{1} $ for $y_{i} \in \hat{Y}_{n}$ where $\textit{softmax}(\mathbf{z}_{i})=(\frac{e^{z_{1}}}{\sum_{j=1}^{n} e^{z_{j}} },\dots, \frac{e^{z_{n}}}{\sum_{j=1}^{n} e^{z_{j}}})$ and $\textit{softmax}(\cdot)_{1}$ is the first element of $\textit{softmax}(\cdot)$.
	It represents the probability of the class in the first index.
	\item  \textbf{Task 2}: All graph representations in a conjunctive word equations are first aggregated by $H_{G}=\frac{1}{n}\sum_{i=1}^{n} H_{G_{i}}$. Then, we compute the score by of each graph by $y_{i}=\textit{softmax}(\text{MLP}_{2}(H_{G_{i}} || H_{G}))_{1}$ for $y_{i} \in \hat{Y}_{n}$ where $\text{MLP}_{2}:\mathbb{R}^{2m} \rightarrow \mathbb{R}^{2}$ is a trainable classifier and $||$ denotes concatenation of two vectors.
	\item  \textbf{Task 3}: We begin by fixing a limit $n$ of equation within a conjunctive word equation. For conjunctive word equations containing more than $n$ word equations, we first sort them by length (in ascending order) 
	and then trim the list to $n$ equations.
	Next, we compute scores for resulting equations using $
	\hat{Y}_{n} = \text{MLP}_{3}(H_{G_{1}}, \dots, H_{G_{n}})$
	where $\text{MLP}_{3}: \mathbb{R}^{nm} \rightarrow \mathbb{R}^{n}$ is a trainable classifier. 
	Scores for any trimmed word equations are set to 0.
	If a conjunctive word equations contains fewer than $n$ word equations, we fill the list with empty equations to reach $n$, 
	and then compute $\hat{Y}_{n}$ in the same way.
\end{itemize}

\subsubsection{Backward Propagation.}
The trainable parameters of the model include the weights of the embedding function 
$\text{MLP}_{0}$, the classifiers $\text{MLP}_{1}$, $\text{MLP}_{2}$, 
$\text{MLP}_{3}$, and the GCNs.
Those trainable parameters are optimized together
by minimizing the categorical cross-entropy loss between the predicted label $\hat{y}_{i}\in\hat{Y}_{n}$ and the true label $y_{i}\in Y_{n}$, using the equation $\mathit{loss}=-\frac{1}{n}\sum_{i=1}^{n} y_{i} \log (\hat{y}_{i})$ where $n$ is the number of conjuncts in the conjunctive word equations.


\subsection{Ranking Options}
\label{section:learning-ranking-options}

In Algorithm~\ref{algorithm:splitEquation}, we introduce seven implementations of \textsc{RankEqs}, aimed at evaluating the efficiency of deterministic versus stochastic ranking methods. 


\begin{itemize}
	\item \textbf{RE1, Baseline:} A baseline defined in Section~\ref{section:split-algorithm}.
	
	\item \textbf{RE2, Random:} \textbf{RE1} is first used to compute the priority of each word equation, and then equations with a priority of 5 are randomly ordered. This approach aims to add some randomness to the baseline.

	
	\item \textbf{RE3, GNN:} Equations ranked at 5 by \textbf{RE1} are then ranked and sorted using the GNN model. While this option incurs higher overhead due to frequent use of the GNN model, it provides the most fine-grained guidance.
	
	\item \textbf{RE4, GNN-Random:} Based on \textbf{RE3}, there is a 50\% chance of invoking the GNN model and a 50\% chance of randomly sorting word equations with a priority of 5. This option provides insight into the performance when introducing a random process into GNN-based ranking.
	
	\item \textbf{RE5, GNN-one-shot:} Based on the priority assigned by \textbf{RE1}, the GNN model is used to rank and sort equations with a priority of 5 the first time they occur, while it is managed by \textbf{RE1} in subsequent iterations. This option invokes the GNN only once to minimize its overhead, while still maintaining its influence on subsequent iterations. Ranking and sorting the word equations early in the process has a greater impact on performance than doing them later.
	
	\item \textbf{RE6, GNN-each-n-iteration:} Based on \textbf{RE3}, 
	instead of calling the GNN model each time multiple word equations have priority 5, it is invoked only every $n = 5000$ calls to the $\textsc{RankEqs}$ function. This option explores a balance between \textbf{RE3} and \textbf{RE5}.
	
	\item \textbf{RE7, GNN-formula-length:} Based on \textbf{RE3}, 
	instead of calling the GNN model each time multiple word equations have priority 5, it is invoked only after $n = 1000$ calls to the $\textsc{RankEqs}$ function when the length of the current word equation does not decrease. This option introduces dynamic control over calling the GNN model.
\end{itemize}

\section{Experimental Results}
\label{section:experiment}

This section describes the benchmarks and the methods used for training data collection. We also compare our evaluation data with leading solvers. The training and prediction workflow is detailed in Appendix~\ref{appendix:workflow}.


\subsection{Benchmarks}
\label{section:experiment-benchmarks}

We initially transformed real-world benchmarks from the non-incremental QF\_S, QF\_SLIA, and QF\_SNLIA tracks of the SMT-LIB benchmark suite~\cite{smtlib:benchmark}, as well as those from the Zaligvinder benchmark suite~\cite{ZaligVinder:benchmark}, into word equation problems by removing length constraints, boolean operators, and regular expressions. However, these transformed problems were overly simplistic, as most solvers, including \textsf{DragonLi}, solved them easily. Consequently, we shifted to evaluating solvers using artificially generated word equation problems inspired by prior research~\cite{day2019solving,10.1007/978-3-031-78709-6_14}.
We summarize the benchmarks as follows:

\begin{itemize}
\item \textbf{Benchmark A1:}
Given a finite set of letters $T$ and a set of variables $V$, the process begins by generating individual word equations of the form $s = s$, where $s$ is a string composed of randomly selected letters from $T$. The maximum length of $s$ is capped at 60.  
Next, substrings in $s$ on both sides of the equation are replaced $n$ times with the concatenation of $m$ fresh variables from $V$.
Here $|T|=6$, $|V|=10$, $n \in [0,5]$, and $m\in [1,5]$. 
Finally, multiple such word equations are conjoined to form a conjunctive word equation problem. The number of equations to be conjoined is randomly selected between 1 and 100.  
Since each replacement variable is a fresh variable from $V$, individual equations in the problem remain linear.

\item \textbf{Benchmark A2:} This benchmark is generated using the same method as Benchmark A1; however, different parameters are employed to increase the difficulty while ensuring that the problem remains linear. Specifically, we use $|T| = 26$, $|V| = 100$, $n \in [0, 16]$, and $m = 1$.

\item \textbf{Benchmark B:} 
This benchmark is generated by the same method as Benchmark A1, except it does not use fresh variables to replace substrings in $s$. This causes a single variable to potentially occur multiple times in an equation, making the problem non-linear.  
The number of equations to be conjoined is randomly picked between 2 and 50, and the maximum length of $s$ is capped at 50.  
In this benchmark, we use $|T| = 10$, $|V| = 10$, $n \in [0, 5]$, and $m = 1$.

\item \textbf{Benchmark C:} We first generate a word equation in the following format:
\begin{equation*}
	\label{eq:track02}
	X_{n}aX_{n}bX_{n-1}\cdots bX_{1} =
	aX_{n}X_{n-1}X_{n-1}b \cdots  X_{1}X_{1}baa~
\end{equation*}
where $X_{1}, ..., X_{n}$ are variables and $a$ and $b$ are letters.
Then, we replace each $b$ with one side of an individual equation generated by Benchmark A1.  
Finally, we join the individual equations to form a conjunctive word equation problem, with the maximum number of conjuncts capped at 100.  
This method ensures that the resulting benchmark is highly non-linear.

\end{itemize}
The statistics of the evaluation data for benchmarks is shown
in Appendix~\ref{appendix:statistics-of-evaluation-data} Table~\ref{tab:eval-data_statistics}. 

%
%
%

\subsection{Training Data Collection}
\label{section:experiment-train-data-collection}
Table~\ref{tab:train-data} outlines the training data collection. We generate 60,000 problems per benchmark and check their satisfiability with \textsf{DragonLi}. For instance, Benchmark A1 contains 1,859 unsolved problems, which are then passed to solvers such as \textsf{Z3}, \textsf{Z3-Noodler}, \textsf{cvc5}, and \textsf{Ostrich}. Together, these solvers identify 181 SAT and 1,678 UNSAT problems, with no single tool able to solve them all.

For UNSAT problems, we extract Minimal Unsatisfiable Subsets (MUSes) using the fastest solver. This yields 909 problems with extractable MUSes, as detailed in Appendix~\ref{appendix:statistics-of-MUS}. We rank word equations within each problem based on their presence in the MUS and their length, then pass the ranked problems back to \textsf{DragonLi}. This allows \textsf{DragonLi} to prioritize word equations appearing in the MUS, enabling it to solve 518 new problems.
Problems in the row \textit{Have MUS} are transformed into a single labeled data (a conjunctive word equation).  
Problems in the row \textit{\textsf{DragonLi} using MUS} are transformed into multiple labeled data, each representing a ranking process step on the shortest path to the solution.

The ranking heuristic’s effectiveness varies with problem benchmarks. For Benchmarks A1 and A2, 57\% to 58\% of problems with MUSes are solved. In Benchmark B, the success rate drops to 20\%, while for Benchmark C, the heuristic has no effect, solving 0 additional problems. Consequently, no training data or model was generated for Benchmark C.
\begin{table}[t]
	\centering
	\caption{Number of problems solved by different solvers and having extracted MUS. The row \textit{Other solvers} shows the number of solved problem in total by \textsf{Z3}, \textsf{Z3-Noodler}, \textsf{cvc5}, and \textsf{Ostrich} where $\checkmark$, $\times$, and $\infty$ denotes SAT, UNSAT, and UKN respectively. The row \textit{\textsf{DragonLi} using MUS} is the number of problems solved by \textsf{DragonLi} when using MUS to rank word equations in the first iteration.}
	\label{tab:train-data}
	\begin{tabular}{cclccclcccclclccclclccc}
		\hline
		Type & \multicolumn{10}{c}{Linear} &  & \multicolumn{11}{c}{Non-linear} \\ \cline{1-11} \cline{13-23} 
		Bench & \multicolumn{5}{c}{A1} &  & \multicolumn{4}{c}{A2} &  & \multicolumn{5}{c}{B} &  & \multicolumn{5}{c}{C} \\ \cline{1-6} \cline{8-11} \cline{13-17} \cline{19-23} 
		Total & \multicolumn{5}{c}{60000} &  & \multicolumn{4}{c}{60000} &  & \multicolumn{5}{c}{60000} &  & \multicolumn{5}{c}{60000} \\ \hline
		\multirow{2}{*}{\textsf{DragonLi}} & \multicolumn{2}{c}{Solved} &  & \multicolumn{2}{c}{$\infty$} &  & Solved &  & \multicolumn{2}{c}{$\infty$} &  & \multicolumn{2}{c}{Solved} &  & \multicolumn{2}{c}{$\infty$} &  & \multicolumn{2}{c}{Solved} &  & \multicolumn{2}{c}{$\infty$} \\ \cline{2-3} \cline{5-6} \cline{8-8} \cline{10-11} \cline{13-14} \cline{16-17} \cline{19-20} \cline{22-23} 
		& \multicolumn{2}{c}{58141} &  & \multicolumn{2}{c}{1859} &  & 50610 &  & \multicolumn{2}{c}{9390} &  & \multicolumn{2}{c}{52056} &  & \multicolumn{2}{c}{7944} &  & \multicolumn{2}{c}{31} &  & \multicolumn{2}{c}{59969} \\ \hline
		\multirow{2}{*}{\begin{tabular}[c]{@{}c@{}}Other\\ solvers\end{tabular}} & \multicolumn{3}{c}{\multirow{2}{*}{}} & $\checkmark$ & $\times$ &  & \multicolumn{2}{c}{\multirow{2}{*}{}} & $\checkmark$ & $\times$ &  & \multicolumn{3}{c}{\multirow{2}{*}{}} & $\checkmark$ & $\times$ &  & \multicolumn{3}{c}{\multirow{2}{*}{}} & $\checkmark$ & $\times$ \\ \cline{5-6} \cline{10-11} \cline{16-17} \cline{22-23} 
		& \multicolumn{3}{c}{} & 181 & 1678 &  & \multicolumn{2}{c}{} & 667 & 4167 &  & \multicolumn{3}{c}{} & 640 & 7304 &  & \multicolumn{3}{c}{} & 383 & 58259 \\ \hline
		Have MUS & \multicolumn{4}{c}{} & 909 &  &  &  &  & 1024 &  & \multicolumn{2}{c}{} &  &  & 2996 &  &  &  &  &  & 15875 \\ \hline
		\begin{tabular}[c]{@{}c@{}}\textsf{DragonLi}\\  using MUS\end{tabular} & \multicolumn{4}{c}{} & 518 &  &  &  &  & 594 &  & \multicolumn{2}{c}{} &  &  & 607 &  &  &  &  &  & 0 \\ \hline
	\end{tabular}
\end{table}



\section{Experimental Settings}

To better investigate the influence of conjuncts order at a conjunctive word equations, we fixed the branch order for all inference rules. Additionally, we fixed the inference rule to the prefix version, meaning it always simplifies the word equation starting from the leftmost term.

Benchmarks were split uniformly into training, validation, and test sets, following standard deep-learning practice. We save the model from the epoch with the highest validation accuracy.

All training records and corresponding hyperparameters, such as a hidden layer size of 128 for all neural networks and number of message passing rounds are available in our repository~\cite{RepositoryReference}.
For example, the experimental results for Benchmark A and Task 2 can be found in~\cite{dragonLi-report-rank}.

Each problem in the benchmarks is evaluated on a computer equipped with two Intel Xeon E5 2630 v4 at 2.20 GHz/core and 128GB memory. The GNNs are trained on NVIDIA A100 GPUs. 
We measured the number of solved problems and the average solving time (in seconds), with timeout of 300 seconds for each proof attempt.

\subsection{Comparison with Other Solvers}
\begin{table}[]
	\caption{Number of problems, average solving time, and average split counts for solvers across four benchmarks. The GNN model used in this table is trained on Task 2. Columns ``UNI", ``CS", and ``CU" indicate uniquely solved, common SAT, and common UNSAT problems, respectively. The ``-" denotes unavailable data. Each benchmark consists of 1000 problems.}
	\label{tab:eval-data}
	\begin{tabular}{|c|c|ccccc|cccc|}
		\hline
		\multirow{2}{*}{Bench} & \multirow{2}{*}{Solver}                                    & \multicolumn{5}{c|}{Number of solved problems}                                                                                                                    & \multicolumn{4}{c|}{\begin{tabular}[c]{@{}c@{}}Average solving time \\ (split number)\end{tabular}}                                                                                                                                                                                                       \\ \cline{3-11} 
		&                                                            & \multicolumn{1}{c|}{SAT}         & \multicolumn{1}{c|}{UNSAT}        & \multicolumn{1}{c|}{UNI} & \multicolumn{1}{c|}{CS}                  & CU                   & \multicolumn{1}{c|}{SAT}                                                      & \multicolumn{1}{c|}{UNSAT}                                                   & \multicolumn{1}{c|}{CS}                                                    & CU                                                            \\ \hline
		\multirow{8}{*}{A1}    & \textsf{DragonLi}                                                   & \multicolumn{1}{c|}{\textbf{24}}          & \multicolumn{1}{c|}{955}          & \multicolumn{1}{c|}{0}   & \multicolumn{1}{c|}{\multirow{8}{*}{13}} & \multirow{8}{*}{678} & \multicolumn{1}{c|}{\begin{tabular}[c]{@{}c@{}}5.6\\ (244.8)\end{tabular}}    & \multicolumn{1}{c|}{\begin{tabular}[c]{@{}c@{}}6.5\\ (1085.3)\end{tabular}}  & \multicolumn{1}{c|}{\begin{tabular}[c]{@{}c@{}}5.0\\ (94.4)\end{tabular}}  & \begin{tabular}[c]{@{}c@{}}5.7\\ (126.3)\end{tabular}         \\ \cline{2-5} \cline{8-11} 
		& \begin{tabular}[c]{@{}c@{}}Random-\\ \textsf{DragonLi}\end{tabular} & \multicolumn{1}{c|}{22}          & \multicolumn{1}{c|}{944}          & \multicolumn{1}{c|}{0}   & \multicolumn{1}{c|}{}                    &                      & \multicolumn{1}{c|}{\begin{tabular}[c]{@{}c@{}}5.6\\ (198.8)\end{tabular}}    & \multicolumn{1}{c|}{\begin{tabular}[c]{@{}c@{}}6.3\\ (932.6)\end{tabular}}   & \multicolumn{1}{c|}{\begin{tabular}[c]{@{}c@{}}5.6\\ (137.6)\end{tabular}} & \begin{tabular}[c]{@{}c@{}}5.7\\ (180.5)\end{tabular}         \\ \cline{2-5} \cline{8-11} 
		& \begin{tabular}[c]{@{}c@{}}GNN-\\ \textsf{DragonLi}\end{tabular}    & \multicolumn{1}{c|}{\textbf{24}} & \multicolumn{1}{c|}{\textbf{961}} & \multicolumn{1}{c|}{0}   & \multicolumn{1}{c|}{}                    &                      & \multicolumn{1}{c|}{\begin{tabular}[c]{@{}c@{}}6.1\\ (164.7)\end{tabular}}    & \multicolumn{1}{c|}{\begin{tabular}[c]{@{}c@{}}7.5\\ (1974.8)\end{tabular}}  & \multicolumn{1}{c|}{\begin{tabular}[c]{@{}c@{}}6.1\\ (96.4)\end{tabular}}  & \begin{tabular}[c]{@{}c@{}}6.3\\ (\textbf{60.5})\end{tabular} \\ \cline{2-5} \cline{8-11} 
		& cvc5                                                       & \multicolumn{1}{c|}{\textbf{24}} & \multicolumn{1}{c|}{952}          & \multicolumn{1}{c|}{1}   & \multicolumn{1}{c|}{}                    &                      & \multicolumn{1}{c|}{0.5}                                                      & \multicolumn{1}{c|}{0.6}                                                     & \multicolumn{1}{c|}{0.1}                                                   & 0.3                                                           \\ \cline{2-5} \cline{8-11} 
		& Z3                                                         & \multicolumn{1}{c|}{17}          & \multicolumn{1}{c|}{960}          & \multicolumn{1}{c|}{0}   & \multicolumn{1}{c|}{}                    &                      & \multicolumn{1}{c|}{8.7}                                                      & \multicolumn{1}{c|}{0.4}                                                     & \multicolumn{1}{c|}{1.1}                                                   & 0.1                                                           \\ \cline{2-5} \cline{8-11} 
		& Z3-Noodler                                                 & \multicolumn{1}{c|}{22}          & \multicolumn{1}{c|}{939}          & \multicolumn{1}{c|}{2}   & \multicolumn{1}{c|}{}                    &                      & \multicolumn{1}{c|}{5.7}                                                      & \multicolumn{1}{c|}{0.3}                                                     & \multicolumn{1}{c|}{4.8}                                                   & 0.1                                                           \\ \cline{2-5} \cline{8-11} 
		& Ostrich                                                    & \multicolumn{1}{c|}{17}          & \multicolumn{1}{c|}{931}          & \multicolumn{1}{c|}{0}   & \multicolumn{1}{c|}{}                    &                      & \multicolumn{1}{c|}{15.0}                                                     & \multicolumn{1}{c|}{5.5}                                                     & \multicolumn{1}{c|}{8.0}                                                   & 4.7                                                           \\ \cline{2-5} \cline{8-11} 
		& Woorpje                                                    & \multicolumn{1}{c|}{23}          & \multicolumn{1}{c|}{744}          & \multicolumn{1}{c|}{0}   & \multicolumn{1}{c|}{}                    &                      & \multicolumn{1}{c|}{3.0}                                                      & \multicolumn{1}{c|}{12.5}                                                    & \multicolumn{1}{c|}{0.1}                                                   & 12.2                                                          \\ \hline
		\multirow{8}{*}{A2}    & \textsf{DragonLi}                                                   & \multicolumn{1}{c|}{59}          & \multicolumn{1}{c|}{824}          & \multicolumn{1}{c|}{0}   & \multicolumn{1}{c|}{\multirow{7}{*}{3}}  & \multirow{7}{*}{0}   & \multicolumn{1}{c|}{\begin{tabular}[c]{@{}c@{}}8.5\\ (4233.4)\end{tabular}}   & \multicolumn{1}{c|}{\begin{tabular}[c]{@{}c@{}}11.8\\ (1231.3)\end{tabular}} & \multicolumn{1}{c|}{\begin{tabular}[c]{@{}c@{}}4.7\\ (27.3)\end{tabular}}  & -                                                             \\ \cline{2-5} \cline{8-11} 
		& \begin{tabular}[c]{@{}c@{}}Random-\\ \textsf{DragonLi}\end{tabular} & \multicolumn{1}{c|}{44}          & \multicolumn{1}{c|}{806}          & \multicolumn{1}{c|}{1}   & \multicolumn{1}{c|}{}                    &                      & \multicolumn{1}{c|}{\begin{tabular}[c]{@{}c@{}}24.7\\ (29779.6)\end{tabular}} & \multicolumn{1}{c|}{\begin{tabular}[c]{@{}c@{}}6.2\\ (210.9)\end{tabular}}   & \multicolumn{1}{c|}{\begin{tabular}[c]{@{}c@{}}4.6\\ (27.3)\end{tabular}}  & -                                                             \\ \cline{2-5} \cline{8-11} 
		& \begin{tabular}[c]{@{}c@{}}GNN-\\ \textsf{DragonLi}\end{tabular}    & \multicolumn{1}{c|}{59}          & \multicolumn{1}{c|}{836}          & \multicolumn{1}{c|}{4}   & \multicolumn{1}{c|}{}                    &                      & \multicolumn{1}{c|}{\begin{tabular}[c]{@{}c@{}}8.4\\ (1330.6)\end{tabular}}   & \multicolumn{1}{c|}{\begin{tabular}[c]{@{}c@{}}11.6\\ (1074.1)\end{tabular}} & \multicolumn{1}{c|}{\begin{tabular}[c]{@{}c@{}}5.9\\ (27.3)\end{tabular}}  & -                                                             \\ \cline{2-5} \cline{8-11} 
		& cvc5                                                       & \multicolumn{1}{c|}{\textbf{67}} & \multicolumn{1}{c|}{142}          & \multicolumn{1}{c|}{15}  & \multicolumn{1}{c|}{}                    &                      & \multicolumn{1}{c|}{0.6}                                                      & \multicolumn{1}{c|}{56.0}                                                    & \multicolumn{1}{c|}{0.1}                                                   & -                                                             \\ \cline{2-5} \cline{8-11} 
		& Z3                                                         & \multicolumn{1}{c|}{8}           & \multicolumn{1}{c|}{\textbf{870}} & \multicolumn{1}{c|}{10}  & \multicolumn{1}{c|}{}                    &                      & \multicolumn{1}{c|}{1.1}                                                      & \multicolumn{1}{c|}{0.6}                                                     & \multicolumn{1}{c|}{0.1}                                                   & -                                                             \\ \cline{2-5} \cline{8-11} 
		& Z3-Noodler                                                 & \multicolumn{1}{c|}{22}          & \multicolumn{1}{c|}{7}            & \multicolumn{1}{c|}{1}   & \multicolumn{1}{c|}{}                    &                      & \multicolumn{1}{c|}{15.4}                                                     & \multicolumn{1}{c|}{3.8}                                                     & \multicolumn{1}{c|}{0.4}                                                   & -                                                             \\ \cline{2-5} \cline{8-11} 
		& Ostrich                                                    & \multicolumn{1}{c|}{13}          & \multicolumn{1}{c|}{18}           & \multicolumn{1}{c|}{2}   & \multicolumn{1}{c|}{}                    &                      & \multicolumn{1}{c|}{24.8}                                                     & \multicolumn{1}{c|}{38.8}                                                    & \multicolumn{1}{c|}{8.6}                                                   & -                                                             \\ \cline{2-11} 
		& Woorpje                                                    & \multicolumn{1}{c|}{0}           & \multicolumn{1}{c|}{0}            & \multicolumn{1}{c|}{0}   & \multicolumn{1}{c|}{-}                   & -                    & \multicolumn{1}{c|}{-}                                                        & \multicolumn{1}{c|}{-}                                                       & \multicolumn{1}{c|}{-}                                                     & -                                                             \\ \hline
		\multirow{8}{*}{B}     & \textsf{DragonLi}                                                   & \multicolumn{1}{c|}{11}          & \multicolumn{1}{c|}{805}          & \multicolumn{1}{c|}{0}   & \multicolumn{1}{c|}{\multirow{8}{*}{4}}  & \multirow{8}{*}{294} & \multicolumn{1}{c|}{\begin{tabular}[c]{@{}c@{}}4.9\\ (62.5)\end{tabular}}     & \multicolumn{1}{c|}{\begin{tabular}[c]{@{}c@{}}5.2\\ (81.5)\end{tabular}}    & \multicolumn{1}{c|}{\begin{tabular}[c]{@{}c@{}}4.9\\ (29.2)\end{tabular}}  & \begin{tabular}[c]{@{}c@{}}5.3\\ (82.4)\end{tabular}          \\ \cline{2-5} \cline{8-11} 
		& \begin{tabular}[c]{@{}c@{}}Random-\\ \textsf{DragonLi}\end{tabular} & \multicolumn{1}{c|}{10}          & \multicolumn{1}{c|}{894}          & \multicolumn{1}{c|}{0}   & \multicolumn{1}{c|}{}                    &                      & \multicolumn{1}{c|}{\begin{tabular}[c]{@{}c@{}}5.0\\ (58.7)\end{tabular}}     & \multicolumn{1}{c|}{\begin{tabular}[c]{@{}c@{}}5.8\\ (295.2)\end{tabular}}   & \multicolumn{1}{c|}{\begin{tabular}[c]{@{}c@{}}5.0\\ (27.25)\end{tabular}} & \begin{tabular}[c]{@{}c@{}}5.2\\ (73.1)\end{tabular}          \\ \cline{2-5} \cline{8-11} 
		& \begin{tabular}[c]{@{}c@{}}GNN-\\ \textsf{DragonLi}\end{tabular}    & \multicolumn{1}{c|}{11}          & \multicolumn{1}{c|}{821}          & \multicolumn{1}{c|}{0}   & \multicolumn{1}{c|}{}                    &                      & \multicolumn{1}{c|}{\begin{tabular}[c]{@{}c@{}}6.5\\ (65.1)\end{tabular}}     & \multicolumn{1}{c|}{\begin{tabular}[c]{@{}c@{}}6.8\\ (70.0)\end{tabular}}    & \multicolumn{1}{c|}{\begin{tabular}[c]{@{}c@{}}6.5\\ (28.25)\end{tabular}} & \begin{tabular}[c]{@{}c@{}}6.8\\ (\textbf{60.2})\end{tabular} \\ \cline{2-5} \cline{8-11} 
		& cvc5                                                       & \multicolumn{1}{c|}{12}          & \multicolumn{1}{c|}{915}          & \multicolumn{1}{c|}{0}   & \multicolumn{1}{c|}{}                    &                      & \multicolumn{1}{c|}{0.1}                                                      & \multicolumn{1}{c|}{0.6}                                                     & \multicolumn{1}{c|}{0.1}                                                   & 0.7                                                           \\ \cline{2-5} \cline{8-11} 
		& Z3                                                         & \multicolumn{1}{c|}{11}          & \multicolumn{1}{c|}{859}          & \multicolumn{1}{c|}{3}   & \multicolumn{1}{c|}{}                    &                      & \multicolumn{1}{c|}{0.1}                                                      & \multicolumn{1}{c|}{0.2}                                                     & \multicolumn{1}{c|}{0.1}                                                   & 0.1                                                           \\ \cline{2-5} \cline{8-11} 
		& Z3-Noodler                                                 & \multicolumn{1}{c|}{\textbf{24}} & \multicolumn{1}{c|}{911}          & \multicolumn{1}{c|}{1}   & \multicolumn{1}{c|}{}                    &                      & \multicolumn{1}{c|}{4.9}                                                      & \multicolumn{1}{c|}{0.4}                                                     & \multicolumn{1}{c|}{1.3}                                                   & 0.4                                                           \\ \cline{2-5} \cline{8-11} 
		& Ostrich                                                    & \multicolumn{1}{c|}{12}          & \multicolumn{1}{c|}{\textbf{917}} & \multicolumn{1}{c|}{2}   & \multicolumn{1}{c|}{}                    &                      & \multicolumn{1}{c|}{6.9}                                                      & \multicolumn{1}{c|}{3.7}                                                     & \multicolumn{1}{c|}{3.3}                                                   & 4.2                                                           \\ \cline{2-5} \cline{8-11} 
		& Woorpje                                                    & \multicolumn{1}{c|}{19}          & \multicolumn{1}{c|}{330}          & \multicolumn{1}{c|}{1}   & \multicolumn{1}{c|}{}                    &                      & \multicolumn{1}{c|}{29.5}                                                     & \multicolumn{1}{c|}{6.0}                                                     & \multicolumn{1}{c|}{0.2}                                                   & 5.0                                                           \\ \hline
		\multirow{8}{*}{C}     & \textsf{DragonLi}                                                   & \multicolumn{1}{c|}{2}           & \multicolumn{1}{c|}{0}            & \multicolumn{1}{c|}{0}   & \multicolumn{1}{c|}{-}                   & -                    & \multicolumn{1}{c|}{\begin{tabular}[c]{@{}c@{}}5.1\\ (85.5)\end{tabular}}     & \multicolumn{1}{c|}{-}                                                       & \multicolumn{1}{c|}{-}                                                     & -                                                             \\ \cline{2-11} 
		& \begin{tabular}[c]{@{}c@{}}Random-\\ \textsf{DragonLi}\end{tabular} & \multicolumn{1}{c|}{2}           & \multicolumn{1}{c|}{0}            & \multicolumn{1}{c|}{0}   & \multicolumn{1}{c|}{-}                   & -                    & \multicolumn{1}{c|}{\begin{tabular}[c]{@{}c@{}}5.0\\ (85.5)\end{tabular}}     & \multicolumn{1}{c|}{-}                                                       & \multicolumn{1}{c|}{-}                                                     & -                                                             \\ \cline{2-11} 
		& \begin{tabular}[c]{@{}c@{}}GNN-\\ \textsf{DragonLi}\end{tabular}    & \multicolumn{1}{c|}{-}           & \multicolumn{1}{c|}{-}            & \multicolumn{1}{c|}{-}   & \multicolumn{1}{c|}{-}                   & -                    & \multicolumn{1}{c|}{-}                                                        & \multicolumn{1}{c|}{-}                                                       & \multicolumn{1}{c|}{-}                                                     & -                                                             \\ \cline{2-11} 
		& cvc5                                                       & \multicolumn{1}{c|}{0}           & \multicolumn{1}{c|}{\textbf{909}} & \multicolumn{1}{c|}{17}  & \multicolumn{1}{c|}{-}                   & \multirow{5}{*}{1}   & \multicolumn{1}{c|}{-}                                                        & \multicolumn{1}{c|}{46.9}                                                    & \multicolumn{1}{c|}{-}                                                     & 17.3                                                          \\ \cline{2-6} \cline{8-11} 
		& Z3                                                         & \multicolumn{1}{c|}{1}           & \multicolumn{1}{c|}{821}          & \multicolumn{1}{c|}{12}  & \multicolumn{1}{c|}{1}                   &                      & \multicolumn{1}{c|}{0.8}                                                      & \multicolumn{1}{c|}{1.7}                                                     & \multicolumn{1}{c|}{0.8}                                                   & 0.1                                                           \\ \cline{2-6} \cline{8-11} 
		& Z3-Noodler                                                 & \multicolumn{1}{c|}{\textbf{7}}  & \multicolumn{1}{c|}{657}          & \multicolumn{1}{c|}{4}   & \multicolumn{1}{c|}{1}                   &                      & \multicolumn{1}{c|}{0.2}                                                      & \multicolumn{1}{c|}{94.1}                                                    & \multicolumn{1}{c|}{0.1}                                                   & 1.0                                                           \\ \cline{2-6} \cline{8-11} 
		& Ostrich                                                    & \multicolumn{1}{c|}{0}           & \multicolumn{1}{c|}{61}           & \multicolumn{1}{c|}{0}   & \multicolumn{1}{c|}{-}                   &                      & \multicolumn{1}{c|}{-}                                                        & \multicolumn{1}{c|}{77.2}                                                    & \multicolumn{1}{c|}{-}                                                     & 27.1                                                          \\ \cline{2-6} \cline{8-11} 
		& Woorpje                                                    & \multicolumn{1}{c|}{3}           & \multicolumn{1}{c|}{62}           & \multicolumn{1}{c|}{0}   & \multicolumn{1}{c|}{1}                   &                      & \multicolumn{1}{c|}{65.0}                                                     & \multicolumn{1}{c|}{28.4}                                                    & \multicolumn{1}{c|}{0.2}                                                   & 223.1                                                         \\ \hline
	\end{tabular}
\end{table}

Table~\ref{tab:eval-data} compares the results of three \textsc{RankEqs} options, \textbf{RE1}, \textbf{RE2}, and \textbf{RE5} (corresponding to \textsf{DragonLi}, Random-\textsf{DragonLi}, and GNN-\textsf{DragonLi}), against five solvers: \textsf{Z3}~\cite{demoura2008z}, \textsf{Z3-Noodler}~\cite{10.1007/978-3-031-57246-3_2}, \textsf{cvc5}~\cite{10.1007/978-3-030-99524-9_24}, \textsf{Ostrich}~\cite{DBLP:journals/pacmpl/ChenHLRW19}, and \textsf{Woopje}~\cite{day2019solving}.  

The primary metric is the number of solved problems.  
In Benchmark A1, GNN-\textsf{DragonLi} achieves the best performance for both SAT and UNSAT problems.  
For Benchmark A2, GNN-\textsf{DragonLi} solves the most problems overall  (895 problems solved), despite not being the best in either category individually.  
GNN-\textsf{DragonLi} outperforms both \textsf{DragonLi} and Random-\textsf{DragonLi}, showing the effectiveness of data-driven heuristics over fixed and random heuristics.  

As problem non-linearity increases (in Benchmark B), some solvers outperform all
\textsf{DragonLi} options.  
For highly non-linear problems (Benchmark C), \textsf{DragonLi} solves almost no problems,
regardless of the options. This is an effect entirely orthogonal to the ranking problem,
however: for non-linear equations, substituting variables that appear multiple times can increase equation length,
resulting in mostly infinite branches in the search tree. 
It then becomes more important to implement additional
criteria to detect unsatisfiable equations,
for instance in terms of word length or
letter count (e.g., \cite{DBLP:conf/fmcad/KumarM21}), which are present in other solvers.
\textsf{DragonLi} deliberately does not include such optimizations, as we aim at investigating the ranking problem in a controlled setting.

For commonly solved problems, the average solving time provides sufficient data only for Benchmarks A1 and B (678 and 294 problems, respectively). In these cases, \textsf{DragonLi} shows no time advantage, partly due to its implementation in \textsf{Python}. Re-implementing the algorithm in a more efficient language, such as Rust~\cite{RepositoryReferenceRust}, can yield over a 100x speedup for single word equation problems.

We also measure the average number of splits in solved problems to evaluate ranking efficiency. GNN-\textsf{DragonLi} demonstrates fewer average splits compared to other options, indicating higher problem-solving efficiency in Benchmarks A1 and B. Our results can be summarized as follows:

\begin{enumerate}
	\item For linear problems, all \textsf{DragonLi} ranking options perform competitively, with GNN-\textsf{DragonLi} solving the highest number of problems.
	\item For moderately non-linear problems (Benchmark B), \textsf{DragonLi} shows moderate performance, but the ranking heuristic offers limited benefits to GNN-\textsf{DragonLi}, leading to reduced performance compared to other options.
	\item For highly non-linear problems (Benchmark C), \textsf{DragonLi} fails to solve most problems due to limitations in its calculus.
	\item The current implementation of \textsf{DragonLi} offers no time advantage for commonly solved problems, though significant improvements are achievable through reimplementation.
\end{enumerate}

Increasing training data for Benchmark A2 from 20,000 to 60,000 allowed GNN-\textsf{DragonLi} to solve additional problems, suggesting that larger training sets may enhance performance.  
An ablation study on alternative \textsc{RankEqs} options is provided in Appendix~\ref{appendix:as}. 
All benchmarks, evaluation results, and implementation details, including hyperparameters, are available in our GitHub repository~\cite{RepositoryReference}.

\section{Related Work}

Axel Thue~\cite{DBLP:journals/corr/Power13} laid the theoretical foundation of word equations by studying the combinatorics of words and sequences, providing an initial understanding of repetitive patterns.
The first deterministic algorithm to solve word equations was proposed by Makanin~\cite{Mak77}, but the complexity is non-elementary. 
Plandowski~\cite{814622} designed an algorithm that reduces the complexity to PSPACE by using a form of run-length encoding to represent strings and variables more compactly during the solving process.
Artur Jeż~\cite{jez2014recompressionsimplepowerfultechnique} proposed a nondeterministic algorithm that runs in $O(n~\text{log}~n)$ space.
Closer to our approach, recent research has focused on improving the practical efficiency of solving word equations.
Perrin and Pin~\cite{pin:hal-00112831} offered an automata-based technique that represents equations in terms of states and transitions. This allows the automata to capture the behavior of strings satisfying the equation.
Markus et al.~\cite{10.1007/978-3-540-24597-1_14} explored graph representations and graph traversal methods to optimize the solving process for word equations, while Day et al.~\cite{day2019solving} reformulated the word equation problem as a reachability problem for nondeterministic finite automata, then encoded it as a propositional satisfiability problem that can be handled by SAT solvers.
Day et al.~\cite{10.1145/3372020.3391556} proposed a transformation system that extends the Nielsen transformation~\cite{levi1944semigroups} to work with linear length constraints.


Deep learning~\cite{GoodBengCour16} has been integrated with various formal verification techniques, such as scheduling SMT solvers~\cite{9643296}, loop invariant reasoning~\cite{NEURIPS201865b1e92c,10.1007/978-3-030-53291-8-9}, and guiding premise selection for Automated Theorem Provers (ATPs)~\cite{Jakubuv2020,10.5555/3294996.3295038}. 
Closely related work in learning from Minimal Unsatisfiable Subsets (MUSes) includes NeuroSAT~\cite{DBLP:journals/corr/abs-1903-04671,selsam2019learning}, which utilizes GNNs to predict the probability of variables appearing in unsat cores, guiding variable branching decisions for Conflict-Driven Clause Learning (CDCL)~\cite{MarquesSilva1999GRASPAS}. 
Additionally, some recent works~\cite{10.1007/978-3-031-50524-9_13,liang2022exploring} explore learning MUSes to guide CHC~\cite{Horn_1951} solvers.

\section{Conclusion and Future Work}
\label{section:conclusion-and-future-work}

In this work, we extend a Nielsen transformation based algorithm~\cite{10.1007/978-3-031-78709-6_14} to support the ranking of conjunctive word equations.
We adapt a multi-classification task to handle a variable number of inputs in three different ways in the ranking task. The model is trained using MUSes to guide the algorithm in solving \UNSAT problems more efficiently. To capture global information in conjunctive word equations, we propose a novel graph representation for word equations. Additionally, we explore various options for integrating the trained model into the algorithms.
Experimental results show that, for linear benchmarks, our framework outperforms the listed leading solvers. However, for non-linear problems, its advantages diminish due to the inherent limitations of the inference rules.
Our framework not only offers a method for ranking word equations but also provides a generalized approach that can be extended to a wide range of formula ranking problems which plays a critical role is symbolic reasoning. 

As future work, we aim to optimize GNN overhead, integrate GNN guidance for both branching and ranking, and extend the solver to support length constraints and regular expressions for greater real-world applicability.
Our framework can be  generalized to handle more decision processes in symbolic methods that take symbolic expressions as input and output a decision choices.

\clearpage

%
%
%
\bibliographystyle{splncs04}
\bibliography{mybibliography}

\clearpage

\appendix

\section{Statistics of Evaluation Data}
\label{appendix:statistics-of-evaluation-data}
Table~\ref{tab:eval-data_statistics} presents the statistics of the evaluation data for each benchmark.  
For each benchmark, we generate a total of 1000 problems for evaluation.  
The total variable and letter occurrence ratios are computed by counting the total occurrences of variables and letters across all problems in a benchmark and dividing by the total number of terms.  
The column ``Single equation" reports the length, number of variables, number of letters, and the maximum occurrences of a single variable for individual equations across all problems in the benchmarks.  
The column ``Problem level" measures the number of equations, number of variables, number of letters, and maximum occurrences of a single variable for each conjunctive word equation in the benchmarks.  
For these columns, the metrics reported are the minimum, maximum, average, and standard deviation values.

For our \textsf{DragonLi} and \textsf{Z3}, the difficulties of these benchmarks are proportional to the total variable occurrence ratio and the length of the equations, as an increase in these values leads to a decrease in the number of solved problems, as shown in Table~\ref{tab:eval-data}.

The number of equations in problems for Benchmark B is set to 50, differing from other benchmarks. This is because its average maximum single variable occurrence is high (53.5), meaning that, on average, a single variable appears 53.5 times across all equations in a problem. Intuitively, this can be interpreted as a formula with more constraints. Consequently, when the number of equations is set to 100, nearly all problems become \UNSAT. When training a data-driven model, the model tends to predict \UNSAT to achieve optimal performance. To mitigate this, the number of equations is reduced to 50.

\begin{table}[]
	\centering
	\caption{Statistics of benchmarks in evaluation data. Min, max, avg, and std denote minimum, maximum, average, and standard deviation value respectively.}
	\label{tab:eval-data_statistics}
\begin{tabular}{ccccccccccccccccc}
	\hline
	Bench &  & \multicolumn{3}{c}{A1} &  & \multicolumn{3}{c}{A2} &  & \multicolumn{3}{c}{B} &  & \multicolumn{3}{c}{C} \\ \cline{1-5} \cline{7-9} \cline{11-13} \cline{15-17} 
	Total &  & \multicolumn{3}{c}{1000} &  & \multicolumn{3}{c}{1000} &  & \multicolumn{3}{c}{1000} &  & \multicolumn{3}{c}{1000} \\ \hline
	\multicolumn{17}{c}{Total variable and letter occurrence ratio} \\ \hline
	Varriable &  & \multicolumn{3}{c}{0.135} &  & \multicolumn{3}{c}{0.130} &  & \multicolumn{3}{c}{0.375} &  & \multicolumn{3}{c}{0.416} \\ \cline{1-5} \cline{7-9} \cline{11-13} \cline{15-17} 
	Letter &  & \multicolumn{3}{c}{0.864} &  & \multicolumn{3}{c}{0.869} &  & \multicolumn{3}{c}{0.625} &  & \multicolumn{3}{c}{0.583} \\ \hline
	\multicolumn{1}{l}{} & \multicolumn{1}{l}{} & \multicolumn{1}{l}{} & \multicolumn{1}{l}{} & \multicolumn{1}{l}{} & \multicolumn{1}{l}{} & \multicolumn{1}{l}{} & \multicolumn{1}{l}{} & \multicolumn{1}{l}{} & \multicolumn{1}{l}{} & \multicolumn{1}{l}{} & \multicolumn{1}{l}{} & \multicolumn{1}{l}{} & \multicolumn{1}{l}{} & \multicolumn{1}{l}{} & \multicolumn{1}{l}{} & \multicolumn{1}{l}{} \\ \hline
	Bench &  & A1 &  & A2 &  & B &  & C &  & A1 &  & A2 &  & B &  & C \\ \cline{1-3} \cline{5-5} \cline{7-7} \cline{9-9} \cline{11-11} \cline{13-13} \cline{15-15} \cline{17-17} 
	&  & \multicolumn{7}{c}{Single equation} &  & \multicolumn{7}{c}{Problem level} \\ \cline{3-9} \cline{11-17} 
	&  & \multicolumn{7}{c}{Length} &  & \multicolumn{7}{c}{Number of equations} \\ \cline{1-9} \cline{11-17} 
	min &  & 2 &  & 74 &  & 2 &  & 13 &  & 1 &  & 1 &  & 2 &  & 1 \\ \cline{1-3} \cline{5-5} \cline{7-7} \cline{9-9} \cline{11-11} \cline{13-13} \cline{15-15} \cline{17-17} 
	max &  & 121 &  & 120 &  & 110 &  & 372 &  & 100 &  & 100 &  & 50 &  & 100 \\ \cline{1-3} \cline{5-5} \cline{7-7} \cline{9-9} \cline{11-11} \cline{13-13} \cline{15-15} \cline{17-17} 
	avg &  & 38.7 &  & 103.2 &  & 43.4 &  & 155.0 &  & 51.4 &  & 50.9 &  & 26.8 &  & 51.8 \\ \cline{1-3} \cline{5-5} \cline{7-7} \cline{9-9} \cline{11-11} \cline{13-13} \cline{15-15} \cline{17-17} 
	std &  & 24.3 &  & 7.2 &  & 20.6 &  & 82.0 &  & 28.8 &  & 28.8 &  & 14.0 &  & 28.6 \\ \cline{1-9} \cline{11-17} 
	&  & \multicolumn{7}{c}{Number of variables} &  & \multicolumn{7}{c}{Number of variables} \\ \cline{1-9} \cline{11-17} 
	min &  & 0 &  & 1 &  & 0 &  & 6 &  & 1 &  & 1 &  & 2 &  & 1 \\ \cline{1-3} \cline{5-5} \cline{7-7} \cline{9-9} \cline{11-11} \cline{13-13} \cline{15-15} \cline{17-17} 
	max &  & 21 &  & 29 &  & 42 &  & 155 &  & 100 &  & 100 &  & 50 &  & 100 \\ \cline{1-3} \cline{5-5} \cline{7-7} \cline{9-9} \cline{11-11} \cline{13-13} \cline{15-15} \cline{17-17} 
	avg &  & 5.2 &  & 13.4 &  & 16.3 &  & 64.5 &  & 51.4 &  & 50.9 &  & 26.8 &  & 51.8 \\ \cline{1-3} \cline{5-5} \cline{7-7} \cline{9-9} \cline{11-11} \cline{13-13} \cline{15-15} \cline{17-17} 
	std &  & 3.0 &  & 4.6 &  & 9.7 &  & 34.1 &  & 28.8 &  & 28.8 &  & 14.0 &  & 28.6 \\ \cline{1-9} \cline{11-17} 
	&  & \multicolumn{7}{c}{Number of letters} &  & \multicolumn{7}{c}{Number of letters} \\ \cline{1-9} \cline{11-17} 
	min &  & 0 &  & 52 &  & 0 &  & 4 &  & 1 &  & 69 &  & 7 &  & 15 \\ \cline{1-3} \cline{5-5} \cline{7-7} \cline{9-9} \cline{11-11} \cline{13-13} \cline{15-15} \cline{17-17} 
	max &  & 120 &  & 119 &  & 100 &  & 264 &  & 3734 &  & 9234 &  & 1796 &  & 10059 \\ \cline{1-3} \cline{5-5} \cline{7-7} \cline{9-9} \cline{11-11} \cline{13-13} \cline{15-15} \cline{17-17} 
	avg &  & 33.4 &  & 89.8 &  & 27.2 &  & 90.5 &  & 1720.4 &  & 4575.1 &  & 729.0 &  & 4689.9 \\ \cline{1-3} \cline{5-5} \cline{7-7} \cline{9-9} \cline{11-11} \cline{13-13} \cline{15-15} \cline{17-17} 
	std &  & 25.1 &  & 11.3 &  & 23.3 &  & 50.8 &  & 975.3 &  & 2593.4 &  & 401.2 &  & 2623.3 \\ \cline{1-9} \cline{11-17} 
	&  & \multicolumn{7}{c}{Max single variable occurrences} &  & \multicolumn{7}{c}{Max single variable occurrences} \\ \cline{1-9} \cline{11-17} 
	min &  & 0 &  & 1 &  & 0 &  & 3 &  & 1 &  & 1 &  & 0 &  & 4 \\ \cline{1-3} \cline{5-5} \cline{7-7} \cline{9-9} \cline{11-11} \cline{13-13} \cline{15-15} \cline{17-17} 
	max &  & 1 &  & 1 &  & 10 &  & 14 &  & 34 &  & 29 &  & 109 &  & 14 \\ \cline{1-3} \cline{5-5} \cline{7-7} \cline{9-9} \cline{11-11} \cline{13-13} \cline{15-15} \cline{17-17} 
	avg &  & 0.97 &  & 1 &  & 3.6 &  & 6.2 &  & 15.9 &  & 13.1 &  & 53.5 &  & 10.2 \\ \cline{1-3} \cline{5-5} \cline{7-7} \cline{9-9} \cline{11-11} \cline{13-13} \cline{15-15} \cline{17-17} 
	std &  & 0.16 &  & 0 &  & 2.1 &  & 1.9 &  & 7.9 &  & 6.0 &  & 26.3 &  & 1.3 \\ \hline
\end{tabular}
\end{table}

\section{Ablation Study}
\label{appendix:as}

For simplicity, we also call a set of conjunctive word equations \textit{a word equation system}.

\subsection{Word Equation Graph Encoding}
\label{appendix:word-equation-graph-encoding}
Before proposing the word equation graph encoding with global information (Figure~\ref{fig:word-equation-graphs}), we initially adopted the graph encoding method introduced in~\cite{10.1007/978-3-031-78709-6_14}. However, this approach yielded low accuracy during training and did not improve solver performance when the trained model was integrated.

\begin{figure}[t]
	\begin{center}
		\includegraphics[width=0.4\textwidth,trim=20 15 25 10]{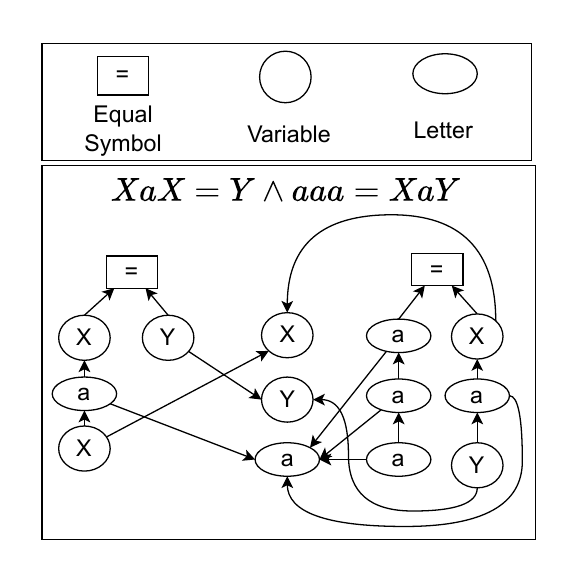}
	\end{center}
	
	\caption{Encode the conjunctive word equations $XaX=Y \wedge aaa=XaY$ to one graph where $X, Y$ are variables and $a$ is a letter.
	}\label{fig:word-equation-global-one-graph}
\end{figure}

We then explored encoding the entire word equation system as a single graph, as illustrated in Figure~\ref{fig:word-equation-global-one-graph}. Several variations were tested, including adding abstract nodes to aggregate all word equations and modifying edge directions. However, these changes did not enhance the final performance in terms of newly solved problems. Difficult instances often involve a large number of equations, making the graph too large and computationally expensive for encoding and GNNs to process. This significantly slows down the solver.

In contrast, using an encoding that represents individual word equations while sharing global information (Figure~\ref{fig:word-equation-graphs}) allows the solver to cache GNN embeddings of unchanged equations during the search. This caching mechanism greatly improves overall efficiency. Such caching is not possible in the design where the entire equation system is encoded as one graph (Figure~\ref{fig:word-equation-global-one-graph}), leading to inferior performance.

\subsection{Training Data Collection}
In Section~\ref{section:training_data_collection}, we mention only two sources of training data. However, our initial data collection included four sources:

\begin{enumerate}
	\item Shortest paths from SAT and UNSAT problems solved by our solver.
	\item MUSes extracted by our solver.
	\item MUSes provided by other solvers.
	\item Shortest paths from UNSAT problems solved by our solver, guided by MUSes from other solvers.
\end{enumerate}

We experimented with all combinations of these sources and ultimately reported only the most effective one in Section~\ref{section:training_data_collection}. The best-performing combination includes MUSes from other solvers and shortest paths from UNSAT problems solved by our solver guided by these MUSes.

\subsection{GNN Model}
We experimented with various GNN architectures, including GCN, Graph Attention Networks (GAT)~\cite{velickovic2018graph}, and Graph Isomorphism Network (GIN)~\cite{xu2019how}. Variants were also tested, such as changing the node aggregation function from mean to max or min. Additionally, we explored combining multiple GNNs by feeding the same input graph into different models, concatenating their outputs, and passing the result to a classifier.

In some cases, particularly when the word equations are long, GAT achieved slightly better validation accuracy during training, with an improvement of about two percentage points. However, when integrating the trained models into the solver, the number of solved problems remained similar to that of GCN.

We attribute this to the higher computational cost of GNNs like GAT and GIN. Although they may offer marginally better performance during training, the gain has limited impact on the final evaluation due to runtime constraints.

This observation held across all tested GNNs and their variants: none outperformed GCN in terms of the number of solved problems. Therefore, we report only the results using GCN in our paper.

\subsection{Ranking Options}
We tested several hand‑crafted ranking options, including ordering equations by size in both ascending and descending order.
Across most benchmarks, these heuristics performed no better, and often worse, than a random order. Consequently, we report only the two simplest baselines: the predefined order (which saves reordering time) and the random order.

\subsection{Experimental Results for Tasks and Integrating Options}

Table~\ref{tab:eval-table-gnn} presents the number of solved \SAT and \UNSAT problems across different training tasks and GNN options for implementing \textsc{RankEqs}. Benchmark C is excluded from evaluation because the ranking process has no significant impact on performance when non-linearity is high, resulting in insufficient data to train the GNN models.

The differences in the number of solved \SAT problems across various configurations are relatively small. This can be attributed to two primary reasons. First, if each conjunct in a conjunctive formula is independent, the ordering would not impact the outcome for \SAT problems, as all conjuncts must independently satisfy the formula. Second, conjuncts in conjunctive word equations are usually not fully independent, as they often share variables. Consequently, the sequence in which the conjuncts are processed influences the solving time for \SAT problems.

Conversely, for \UNSAT problems, the sequence of the conjuncts affects performance more regardless of the independence of individual conjuncts.

Therefore, the following discussion primarily focuses on the \UNSAT problems.

In terms of the training tasks, the computational overhead follows the order $\text{Task}~2 > \text{Task}~1 > \text{Task}~3$. The reasons are outlined as follows:

\begin{itemize}
	\item \textbf{Task~2}: This task first computes the graph representation for each word equation, $H_{G_{i}}$, and then aggregates these into a global feature representation using $H_{G} = \sum (H_{G_{1}}, \dots, H_{G_{n}})$. To compute the ranking score of each conjunct in the conjunctive word equations, forward propagation must be performed $n$ times with the concatenated input $H_{G_{i}} || H_{G}$ for each GNN layer.
	\item \textbf{Task~1}: This task requires forward propagation $n$ times using the individual graph representations $H_{G_{i}}$, but it does not involve computing or concatenating the global feature representation $H_{G}$.
	\item \textbf{Task~3}: This task only requires a single forward propagation step using the set of individual graph representations $(H_{G_{1}}, \dots, H_{G_{n}})$.
\end{itemize}

The numbers of solved \UNSAT problems in columns \textbf{RE3} and \textbf{RE4} for Benchmarks A2 and B provide supporting evidence. 
For Benchmark A1, it is not sensitive to the overhead of GNN calls because the overall number of iterations required to solve the problems is low.

In terms of GNN options, \textbf{RE3} involves the most of calls to the GNN model. 
It is generally outperformed by other options across all tasks and benchmarks due to its overhead.

\textbf{RE4} involves a random process. Its performance is below average for Benchmarks A1 and A2 because it disrupts the predictions of GNNs. 
For Benchmark B, it achieves the best performance for Tasks 1 and 3. This is because Benchmark B is non-linear; repeatedly applying inference rules to the same non-linear word equation leads to an increase in length and potentially non-termination. The random process helps to escape such situations. 
However, for Task 3, the overhead of calling the GNN diminishes its advantage.

\textbf{RE5} has the least overhead from model calls. For Benchmarks A1 and A2, it outperforms most other options. For Benchmark B, it achieves average performance.

\textbf{RE6} and \textbf{RE7} maintain high performance across all benchmarks but do not consistently outperform \textbf{RE5}.

\begin{table}[]
	\centering
	\caption{Number of solved problems for different GNN options and training tasks. The bold numbers, along with their corresponding GNN options and tasks, are referenced in the rows labeled ``GNN'' in Table~\ref{tab:eval-data}. The columns ``3'' ,``4", ..., ``7" corresponds to \textbf{RE3}, \textbf{RE4}, ...,  \textbf{RE7} in Section~\ref{section:learning-ranking-options} respectively.
	 }
	\label{tab:eval-table-gnn}
	\begin{tabular}{|cc|ccccccccccccccc|}
		\hline
		\multicolumn{2}{|c|}{Bench}                      & \multicolumn{5}{c|}{A1}                                                                                                              & \multicolumn{5}{c|}{A2}                                                                                                              & \multicolumn{5}{c|}{B}                                                                                          \\ \hline
		\multicolumn{2}{|c|}{}                           & \multicolumn{15}{c|}{$\textsc{RankEqs}$ GNN options}                                                                                                                                                                                                                                                                                                                                          \\ \hline
		\multicolumn{1}{|c|}{Tasks}              &       & \multicolumn{1}{c|}{3}   & \multicolumn{1}{c|}{4}   & \multicolumn{1}{c|}{5}   & \multicolumn{1}{c|}{6}   & \multicolumn{1}{c|}{7}   & \multicolumn{1}{c|}{3}   & \multicolumn{1}{c|}{4}   & \multicolumn{1}{c|}{5}   & \multicolumn{1}{c|}{6}   & \multicolumn{1}{c|}{7}   & \multicolumn{1}{c|}{3}   & \multicolumn{1}{c|}{4}   & \multicolumn{1}{c|}{5}   & \multicolumn{1}{c|}{6}   & 7   \\ \hline
		\multicolumn{1}{|c|}{\multirow{2}{*}{1}} & SAT   & \multicolumn{1}{c|}{23}  & \multicolumn{1}{c|}{23}  & \multicolumn{1}{c|}{24}  & \multicolumn{1}{c|}{24}  & \multicolumn{1}{c|}{24}  & \multicolumn{1}{c|}{34}  & \multicolumn{1}{c|}{42}  & \multicolumn{1}{c|}{59}  & \multicolumn{1}{c|}{59}  & \multicolumn{1}{c|}{59}  & \multicolumn{1}{c|}{10}  & \multicolumn{1}{c|}{10}  & \multicolumn{1}{c|}{11}  & \multicolumn{1}{c|}{11}  & 11  \\ \cline{2-17} 
		\multicolumn{1}{|c|}{}                   & UNSAT & \multicolumn{1}{c|}{945} & \multicolumn{1}{c|}{938} & \multicolumn{1}{c|}{954} & \multicolumn{1}{c|}{955} & \multicolumn{1}{c|}{955} & \multicolumn{1}{c|}{404} & \multicolumn{1}{c|}{521} & \multicolumn{1}{c|}{837} & \multicolumn{1}{c|}{829} & \multicolumn{1}{c|}{824} & \multicolumn{1}{c|}{776} & \multicolumn{1}{c|}{876} & \multicolumn{1}{c|}{815} & \multicolumn{1}{c|}{805} & 833 \\ \hline
		\multicolumn{1}{|c|}{\multirow{2}{*}{2}} & SAT   & \multicolumn{1}{c|}{24}  & \multicolumn{1}{c|}{23}  & \multicolumn{1}{c|}{\textbf{24}}  & \multicolumn{1}{c|}{24}  & \multicolumn{1}{c|}{24}  & \multicolumn{1}{c|}{49}  & \multicolumn{1}{c|}{48}  & \multicolumn{1}{c|}{\textbf{59}}  & \multicolumn{1}{c|}{59}  & \multicolumn{1}{c|}{59}  & \multicolumn{1}{c|}{9}   & \multicolumn{1}{c|}{10}  & \multicolumn{1}{c|}{\textbf{11}}  & \multicolumn{1}{c|}{9}   & 11  \\ \cline{2-17} 
		\multicolumn{1}{|c|}{}                   & UNSAT & \multicolumn{1}{c|}{946} & \multicolumn{1}{c|}{945} & \multicolumn{1}{c|}{\textbf{961}} & \multicolumn{1}{c|}{955} & \multicolumn{1}{c|}{954} & \multicolumn{1}{c|}{300} & \multicolumn{1}{c|}{468} & \multicolumn{1}{c|}{\textbf{836}} & \multicolumn{1}{c|}{833} & \multicolumn{1}{c|}{826} & \multicolumn{1}{c|}{78}  & \multicolumn{1}{c|}{256} & \multicolumn{1}{c|}{\textbf{821}} & \multicolumn{1}{c|}{767} & 829 \\ \hline
		\multicolumn{1}{|c|}{\multirow{2}{*}{3}} & SAT   & \multicolumn{1}{c|}{24}  & \multicolumn{1}{c|}{23}  & \multicolumn{1}{c|}{24}  & \multicolumn{1}{c|}{24}  & \multicolumn{1}{c|}{24}  & \multicolumn{1}{c|}{48}  & \multicolumn{1}{c|}{38}  & \multicolumn{1}{c|}{59}  & \multicolumn{1}{c|}{59}  & \multicolumn{1}{c|}{59}  & \multicolumn{1}{c|}{9}   & \multicolumn{1}{c|}{10}  & \multicolumn{1}{c|}{11}  & \multicolumn{1}{c|}{11}  & 11  \\ \cline{2-17} 
		\multicolumn{1}{|c|}{}                   & UNSAT & \multicolumn{1}{c|}{936} & \multicolumn{1}{c|}{940} & \multicolumn{1}{c|}{951} & \multicolumn{1}{c|}{955} & \multicolumn{1}{c|}{953} & \multicolumn{1}{c|}{825} & \multicolumn{1}{c|}{779} & \multicolumn{1}{c|}{816} & \multicolumn{1}{c|}{832} & \multicolumn{1}{c|}{825} & \multicolumn{1}{c|}{884} & \multicolumn{1}{c|}{892} & \multicolumn{1}{c|}{792} & \multicolumn{1}{c|}{805} & 829 \\ \hline
	\end{tabular}
\end{table}

\section{Statistics of MUSes}
\label{appendix:statistics-of-MUS}
\begin{table}[]
	\centering
	\caption{Statistics of MUSes.}
	\label{tab:MUS-statistics}
\begin{tabular}{ccccccccccccccccc}
	\hline
	Bench &  & \multicolumn{3}{c}{A1} &  & \multicolumn{3}{c}{A2} &  & \multicolumn{3}{c}{B} &  & \multicolumn{3}{c}{C} \\ \cline{1-5} \cline{7-9} \cline{11-13} \cline{15-17} 
	Total &  & \multicolumn{3}{c}{909} &  & \multicolumn{3}{c}{1024} &  & \multicolumn{3}{c}{2996} &  & \multicolumn{3}{c}{15875} \\ \hline
	\multicolumn{17}{c}{Total variable and letter occurrence ratio} \\ \hline
	Varriable &  & \multicolumn{3}{c}{0.124} &  & \multicolumn{3}{c}{0.139} &  & \multicolumn{3}{c}{0.489} &  & \multicolumn{3}{c}{0.408} \\ \cline{1-5} \cline{7-9} \cline{11-13} \cline{15-17} 
	Letter &  & \multicolumn{3}{c}{0.875} &  & \multicolumn{3}{c}{0.861} &  & \multicolumn{3}{c}{0.511} &  & \multicolumn{3}{c}{0.592} \\ \hline
	\multicolumn{1}{l}{} & \multicolumn{1}{l}{} & \multicolumn{1}{l}{} & \multicolumn{1}{l}{} & \multicolumn{1}{l}{} & \multicolumn{1}{l}{} & \multicolumn{1}{l}{} & \multicolumn{1}{l}{} & \multicolumn{1}{l}{} & \multicolumn{1}{l}{} & \multicolumn{1}{l}{} & \multicolumn{1}{l}{} & \multicolumn{1}{l}{} & \multicolumn{1}{l}{} & \multicolumn{1}{l}{} & \multicolumn{1}{l}{} & \multicolumn{1}{l}{} \\ \hline
	Bench &  & A1 &  & A2 &  & B &  & C &  & A1 &  & A2 &  & B &  & C \\ \cline{1-3} \cline{5-5} \cline{7-7} \cline{9-9} \cline{11-11} \cline{13-13} \cline{15-15} \cline{17-17} 
	&  & \multicolumn{7}{c}{Single equation} &  & \multicolumn{7}{c}{Problem level} \\ \cline{3-9} \cline{11-17} 
	&  & \multicolumn{7}{c}{Length} &  & \multicolumn{7}{c}{Number of equations} \\ \cline{1-9} \cline{11-17} 
	min &  & 2 &  & 76 &  & 11 &  & 12 &  & 2 &  & 1 &  & 1 &  & 1 \\ \cline{1-3} \cline{5-5} \cline{7-7} \cline{9-9} \cline{11-11} \cline{13-13} \cline{15-15} \cline{17-17} 
	max &  & 117 &  & 120 &  & 114 &  & 354 &  & 8 &  & 8 &  & 8 &  & 7 \\ \cline{1-3} \cline{5-5} \cline{7-7} \cline{9-9} \cline{11-11} \cline{13-13} \cline{15-15} \cline{17-17} 
	avg &  & 40.6 &  & 101.7 &  & 40.1 &  & 66.2 &  & 2.2 &  & 3.3 &  & 2.6 &  & 2.0 \\ \cline{1-3} \cline{5-5} \cline{7-7} \cline{9-9} \cline{11-11} \cline{13-13} \cline{15-15} \cline{17-17} 
	std &  & 23.5 &  & 7.4 &  & 17.1 &  & 60.8 &  & 0.7 &  & 1.0 &  & 1.3 &  & 0.5 \\ \cline{1-9} \cline{11-17} 
	&  & \multicolumn{7}{c}{Number of variables} &  & \multicolumn{7}{c}{Number of variables} \\ \cline{1-9} \cline{11-17} 
	min &  & 1 &  & 2 &  & 10 &  & 6 &  & 2 &  & 10 &  & 10 &  & 7 \\ \cline{1-3} \cline{5-5} \cline{7-7} \cline{9-9} \cline{11-11} \cline{13-13} \cline{15-15} \cline{17-17} 
	max &  & 16 &  & 26 &  & 41 &  & 148 &  & 49 &  & 117 &  & 173 &  & 525 \\ \cline{1-3} \cline{5-5} \cline{7-7} \cline{9-9} \cline{11-11} \cline{13-13} \cline{15-15} \cline{17-17} 
	avg &  & 5.1 &  & 14.2 &  & 19.6 &  & 27.0 &  & 11.3 &  & 46.3 &  & 50.4 &  & 54.7 \\ \cline{1-3} \cline{5-5} \cline{7-7} \cline{9-9} \cline{11-11} \cline{13-13} \cline{15-15} \cline{17-17} 
	std &  & 2.6 &  & 4.6 &  & 6.4 &  & 25.5 &  & 5.4 &  & 17.3 &  & 26.1 &  & 56.3 \\ \cline{1-9} \cline{11-17} 
	&  & \multicolumn{7}{c}{Number of letters} &  & \multicolumn{7}{c}{Number of letters} \\ \cline{1-9} \cline{11-17} 
	min &  & 1 &  & 57 &  & 1 &  & 4 &  & 4 &  & 69 &  & 1 &  & 4 \\ \cline{1-3} \cline{5-5} \cline{7-7} \cline{9-9} \cline{11-11} \cline{13-13} \cline{15-15} \cline{17-17} 
	max &  & 111 &  & 118 &  & 99 &  & 248 &  & 318 &  & 710 &  & 362 &  & 758 \\ \cline{1-3} \cline{5-5} \cline{7-7} \cline{9-9} \cline{11-11} \cline{13-13} \cline{15-15} \cline{17-17} 
	avg &  & 35.6 &  & 87.5 &  & 20.5 &  & 39.2 &  & 79.2 &  & 258.6 &  & 52.6 &  & 79.3 \\ \cline{1-3} \cline{5-5} \cline{7-7} \cline{9-9} \cline{11-11} \cline{13-13} \cline{15-15} \cline{17-17} 
	std &  & 23.8 &  & 11.4 &  & 17.5 &  & 36.8 &  & 39.0 &  & 90.2 &  & 40.1 &  & 78.7 \\ \cline{1-9} \cline{11-17} 
	&  & \multicolumn{7}{c}{Max single variable occurrences} &  & \multicolumn{7}{c}{Max single variable occurrences} \\ \cline{1-9} \cline{11-17} 
	min &  & 1 &  & 1 &  & 1 &  & 3 &  & 1 &  & 1 &  & 2 &  & 3 \\ \cline{1-3} \cline{5-5} \cline{7-7} \cline{9-9} \cline{11-11} \cline{13-13} \cline{15-15} \cline{17-17} 
	max &  & 1 &  & 1 &  & 10 &  & 14 &  & 1 &  & 1 &  & 10 &  & 14 \\ \cline{1-3} \cline{5-5} \cline{7-7} \cline{9-9} \cline{11-11} \cline{13-13} \cline{15-15} \cline{17-17} 
	avg &  & 1 &  & 1 &  & 4.4 &  & 4.3 &  & 1 &  & 1 &  & 5.4 &  & 4.6 \\ \cline{1-3} \cline{5-5} \cline{7-7} \cline{9-9} \cline{11-11} \cline{13-13} \cline{15-15} \cline{17-17} 
	std &  & 0 &  & 0 &  & 1.5 &  & 1.5 &  & 0 &  & 0 &  & 1.4 &  & 1.6 \\ \hline
\end{tabular}
\end{table}

Table~\ref{tab:MUS-statistics} presents the statistics of equations in the MUS for each benchmark, corresponding to the row ``Have MUS'' in Table~\ref{tab:train-data}. Each conjunctive word equation is associated with one MUS.

\section{Relation to Paper~\cite{10.1007/978-3-031-78709-6_14}}
\label{appendix:relation-to-previous-paper}

Our work is initially inspired by Paper~\cite{10.1007/978-3-031-78709-6_14}, but ultimately, the two studies share only the calculus.
Other parts, such as 
the syntax tree of formula, GNNs, and neural network based embedding and classifier are standard concepts and components applicable to many tasks.
Their differences are summarized as following:

\begin{enumerate}
	\item \textbf{Goal}
	\begin{itemize}
		\item \textbf{Paper~\cite{10.1007/978-3-031-78709-6_14}}: Guides the branching process, which is crucial for solving individual SAT word equations.
		\item \textbf{This paper}: Guides the ranking process before branching, which significantly impacts solving UNSAT word equation systems.
	\end{itemize}
	\item \textbf{Training Data Collection}
	\begin{itemize}
		\item \textbf{Paper~\cite{10.1007/978-3-031-78709-6_14}}: Uses shortest paths from SAT problems.
		\item \textbf{This paper}: Uses MUSes provided by other solvers and shortest paths from UNSAT problems.
	\end{itemize}
	
	\item \textbf{Word Equation Graph Encoding}
	\begin{itemize}
		\item \textbf{Paper~\cite{10.1007/978-3-031-78709-6_14}}: Encodes single word equations based on their syntax trees.
		\item \textbf{This paper}: Encodes single word equations with syntax trees enhanced by additional global information.
	\end{itemize}
	
	\item \textbf{Training Task (Model Structure)}
	\begin{itemize}
		\item \textbf{Paper~\cite{10.1007/978-3-031-78709-6_14}}: Performs a multi-class classification task on single word equation embeddings.
		\item \textbf{This paper}: Performs three different classification tasks on embeddings of word equation systems.
	\end{itemize}
	
	\item \textbf{Integration into the Solver}
	\begin{itemize}
		\item \textbf{Paper~\cite{10.1007/978-3-031-78709-6_14}}: Uses predictions directly or in combination with random branching.
		\item \textbf{This paper}: Uses predictions at varying frequencies, controlled by manually defined criteria, including a random mechanism.
	\end{itemize}
\end{enumerate}

In summary, Paper~\cite{10.1007/978-3-031-78709-6_14} addresses branching decisions, formulated as a simpler classification problem, while our work focuses on equation ordering, which requires solving a ranking problem.

Although both approaches utilize similar neural architectures, their training methodologies differ fundamentally. Classification models optimize a loss function over individual examples, selecting one label from a fixed set. In contrast, ranking models must compare items relative to one another, often requiring pairwise or listwise training, which scales combinatorially with the number of candidates. This makes ranking intrinsically more complex, as it models dependencies between examples rather than treating them in isolation.

Combining these two approaches is non-trivial and is left for future work. From an empirical perspective, isolating each learning objective is essential when studying deep learning-based solvers. Otherwise, the configuration space becomes unmanageable.

\section{Calculus for Word Equations~\cite{10.1007/978-3-031-78709-6_14}}
\label{app:split-rules} 
The calculus for word equations proposed by~\cite{10.1007/978-3-031-78709-6_14} comprises a set of inference rules. Each inference rule is expressed in the following form:
\begin{center}
	\def\arraystretch{2}
	\begin{tabularx}{8cm}{lX|X|X}
		\multirow{2}{*}{$\mathit{Name}$} & \multicolumn{3}{c}{$P$} \\ \cline{2-4}
		& \makecell{$[\mathit{cond_1}]$\\$C_1$} & \makecell{\dots} & \makecell{$[\mathit{cond_n}]$\\$C_n$} 
	\end{tabularx}
\end{center}
Here, $\mathit{Name}$ is the name of the rule, $P$ is the premise, and $C_i$s are the conclusions.
Each $\mathit{cond_i}$ is a substitution that is applied implicitly to the corresponding conclusion~$C_i$, describing the case handled by that particular branch.
In our case, $P$ is a conjunctive word equation and each $C_i$ is either a conjunctive word equation or a final state, \SAT or \UNSAT.

To introduce the inference rules, we use distinct letters~$a, b \in \Sigma$ and variables $X, Y \in \Gamma$, while $u$ and $v$ denote sequences of letters and variables.

Rules $R_{1}$, $R_{2}, R_{3}$, and $R_{4}$ (Figure~\ref{fig:rules_CNF}) define how to conclude SAT, and how to handle equations in which one side is empty. In $R_{3}$, note that the substitution~$X \mapsto \epsilon$ is applied to the conclusion~$\phi$.
Rules $R_{5}$ and $R_{6}$ (Figure~\ref{fig:rules_Rcc}) refer to cases in which each word starts with a letter. The rules simplify the leftmost equation, either by removing the first letter, if it is identical on both sides ($R_{5}$), or by concluding that the equation is \UNSAT ($R_{6}$).
Rule $R_{7}$ (Figure~\ref{fig:rules_Rvc}) manages cases where one side begins with a letter and the other one with a variable. The rule introduces two branches, since the variable must either denote the empty string~$\epsilon$, or its value must start with the same letter as the right-hand side.
Rule $R_{8}$ and $R_{9}$ (Figure~\ref{fig:rules_Rvv}) handles the case in which both sides of an equation start with a variable, implying that either both variables have the same value or the value of one is included in the value of the other.

We implicitly assume symmetric versions of the rules $R_{4}$, $R_{5}$, and $R_{8}$, swapping left-hand side and right-hand side of the equation that is rewritten. For instance, the symmetric rule for $R_{4}$ would have premise $\epsilon = X \wedge \phi$.

We implicitly assume the suffix versions of rules $R_{5}$ to $R_{9}$, operating from the last terms on both sides. For example, the suffix version of $R_{7}$ is
\begin{equation*}
	\begin{tabularx}{5cm}{lX|X}
		\multirow{2}{*}{$R^{\text{suffix}}_{7}$} & \multicolumn{2}{c}{$u \cdot X = v \cdot a \land \phi$} \\ \cline{2-3}
		& \makecell{$[X \mapsto \epsilon]$                         \\$u = v \cdot a \land \phi$} & \makecell{$[X \mapsto  X' \cdot a]
			$\\$u \cdot X'= v \land \phi$}
	\end{tabularx}
\end{equation*}

\begin{figure}[t]
	\fbox{\parbox{0.975\linewidth}{
%
%
%
%

			\begin{minipage}{\textwidth}
				\begin{minipage}{0.2\linewidth}
					\begin{prooftree}
						\rootAtTop
						\AxiomC{\SAT}
						\LeftLabel{$R_1$}
						\UnaryInfC{$True$}
					\end{prooftree}
				\end{minipage}%
				\begin{minipage}{0.25\linewidth}
					\begin{prooftree}
						\rootAtTop
						\AxiomC{$\phi$}
						\LeftLabel{$R_2$}
						\UnaryInfC{$\epsilon=\epsilon \wedge \phi$}
					\end{prooftree}
				\end{minipage}%
				\begin{minipage}{0.25\linewidth}
					\vspace{8pt}
					\begin{prooftree}
						\rootAtTop
						\AxiomC{$\phi$}
						\noLine
						\UnaryInfC{$[X \mapsto \epsilon]$}
						\LeftLabel{$R_3$}
						\UnaryInfC{$X = \epsilon \wedge \phi$}
					\end{prooftree}
				\end{minipage}%
				\begin{minipage}{0.25\linewidth}
					\begin{prooftree}
						\rootAtTop
						\AxiomC{\UNSAT}
						\LeftLabel{$R_4$}
						\UnaryInfC{$a\cdot u=\epsilon \wedge \phi$}
					\end{prooftree}
				\end{minipage}%

				\vspace{0.2cm}
				\centering
				with $X\in \Gamma$ and $a \in \Sigma$.
				
				\subcaption{Simplification Rule}
				\label{fig:rules_CNF}
			\end{minipage}

			\begin{minipage}{\textwidth}
				
				\begin{minipage}{0.50\linewidth}
					\begin{prooftree}
						\rootAtTop
						\AxiomC{\shortstack{$u = v \land \phi$}}
						\LeftLabel{$R_5$}
						\UnaryInfC{$a \cdot u = a \cdot v \land \phi$}
					\end{prooftree}
				\end{minipage}%
				\begin{minipage}{0.50\linewidth}
					\begin{prooftree}
						\rootAtTop
						\AxiomC{\shortstack{\UNSAT}}
						\LeftLabel{$R_6$}
						\UnaryInfC{$a \cdot u = b \cdot v \land \phi$}
					\end{prooftree}
				\end{minipage}%
				
				\vspace{0.2cm}
				\centering
				with $a, b$ two different Letter from $\Sigma$.
				\subcaption{Letter-Letter Rule}
				\label{fig:rules_Rcc}
			\end{minipage}
			
			\vspace{0.5cm}
			
			\begin{minipage}{\textwidth}
				\begin{minipage}{\linewidth}
					\centering
					\begin{tabularx}{5cm}{lX|X}
						\multirow{2}{*}{$R_7$} & \multicolumn{2}{c}{$X \cdot u = a \cdot v \land \phi$} \\ \cline{2-3}
						& \makecell{$[X \mapsto \epsilon]$                         \\$u = a \cdot v \land \phi$} & \makecell{$[X \mapsto a \cdot X']$\\$X'\cdot u = v \land \phi$}
					\end{tabularx}
				\end{minipage}\hfill

				\vspace{0.2cm}
				\centering
				with $X'$ a \emph{fresh} element of $\Gamma$.
				
				\subcaption{Variable-Letter Rules}
				\label{fig:rules_Rvc}
			\end{minipage}
			
			\vspace{0.5cm}
			
			\begin{minipage}{\textwidth}
				
				\begin{minipage}{\linewidth}
					\centering
				\begin{tabularx}{8cm}{lX|X|X}
					\multirow{2}{*}{$R_{8}$} & \multicolumn{3}{c}{$X \cdot u = Y \cdot v \land \phi$} \\ \cline{2-4}
					& \makecell{$[X \mapsto Y]$\\$u = v \land \phi$}  & \makecell{$[X \mapsto Y \cdot Y']$\\$Y'\cdot u = v \land \phi$} & \makecell{$[Y \mapsto X \cdot X']$\\$ u = X' \cdot v \land \phi$} 
				\end{tabularx}
				\end{minipage}
				
				\vspace{0.2cm}
				
				\begin{minipage}{\linewidth}
					\centering
				\begin{tabularx}{3cm}{lX}
					\multirow{2}{*}{$R_{9}$} & $X \cdot u = X \cdot v \land \phi$ \\ \cline{2-2}
					& \makecell{$u =  v \land \phi$ \\ $~$} 
				\end{tabularx}
				\end{minipage}
				
				\centering
				with $X \neq Y$ and $X', Y'$ \emph{fresh} elements of $\Gamma$.
				\subcaption{Variable-Variable Rule}
				\label{fig:rules_Rvv}
				
			\end{minipage}
	}}
	\caption{Inference rules of the proof system for word equations}
	\label{fig:split_rules}
\end{figure}

\section{Workflow}
\label{appendix:workflow}
\begin{figure}[t]
	
	\begin{center}
			\includegraphics[width=\textwidth]{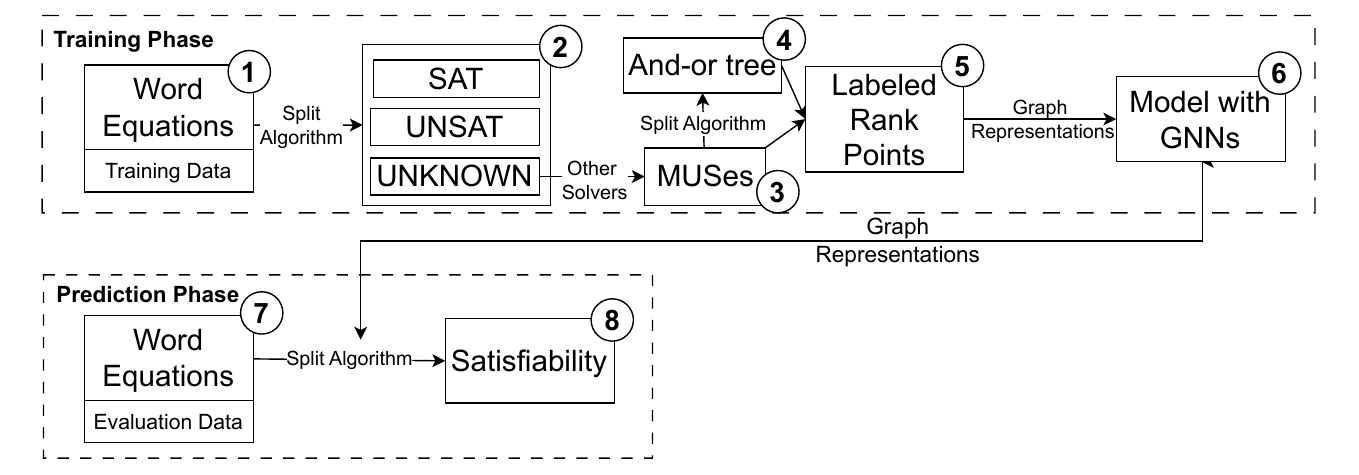}
		\end{center}
	
	\caption{The workflow diagram for the training and prediction phase.}
	\label{figure:workflow}
\end{figure}

The workflow of our framework is illustrated in Figure~\ref{figure:workflow}. For a benchmark, we begin by randomly splitting the dataset into training and evaluation subsets.
During the training phase, we first input the word equation problems into the split algorithm ranking option \textbf{RE1} to obtain their satisfiabilities and separate them into \SAT, \UNSAT, and \UNKNOWN sets. For the \SAT and \UNSAT sets, we discard them since our algorithm already know how to solve them.
For the \UNKNOWN set, the problems are passed to other solvers, such as \textsf{z3} and \textsf{cvc5}. If the solver concludes \UNSAT, we systematically identify Minimal Unsatisfiable Subsets (MUSes) of conjunctive word equations by exhaustively checking the satisfiability of subsets, starting with individual equations and stopping upon finding the first MUS.
We use the MUSes to rank and sort conjunctive word equations unsolvable by the split algorithm, then reprocess the sorted equations with it. This allows the split algorithm to solve some problems and construct proof trees.
Then, we can extract the labeled data from both and-or tree and MUSes. The way of label them can be found in Section \ref{section:learning}.

Next, we convert the labeled conjunctive word equations from textual to graph format, enabling the model with GNN layers to process them. The model takes a rank point (i.e., a conjunction of word equations) as input and outputs corresponding scores that indicate the priority of the conjuncts.

In the prediction phase, during the step where inference rules are applied, the trained model ranks and sorts the conjuncts. The equation with the highest score is then selected for apply the branching inference rules.

\end{document}